%% file: main.tex
\documentclass{article}

\usepackage[preprint]{neurips_2026}

\usepackage[utf8]{inputenc}
\usepackage[T1]{fontenc}

\usepackage[hidelinks]{hyperref}
\usepackage{url}
\usepackage{booktabs}
\usepackage{longtable}
\usepackage{amsfonts}
\usepackage{amsmath}
\usepackage{amssymb}
\usepackage{nicefrac}
\usepackage{microtype}
\usepackage{graphicx}
\usepackage{multirow}
\usepackage{enumitem}
\usepackage{xcolor}
\usepackage{placeins}
\title{VLAQuantBench: Closed-Loop Evaluation of Post-Training Quantization for Vision-Language-Action Models}

\author{Jiuyi Xu$^1$ \quad Qing Jin$^2$ \quad Meida Chen$^3$ \quad Song Wang$^4$ \\
\textbf{Yang Sui}$^5$ \quad \textbf{Yangming Shi}$^1$ \\
$^1$Colorado School of Mines \quad $^2$Independent Researcher \\
$^3$University of Central Florida \quad $^4$USC Institute for Creative Technologies \\
$^5$Microsoft AI}

\begin{document}
\maketitle

\input{tex_files/abstract}
\input{tex_files/introduction}
\input{tex_files/related_work}
\input{tex_files/protocol}
\input{tex_files/experiments}
\input{tex_files/discussion}
\input{tex_files/limitation}
\input{tex_files/conclusion}

\FloatBarrier
\bibliographystyle{plainnat}
\bibliography{custom}

\clearpage
\appendix
\input{tex_files/appendix}

\end{document}

%% file: tex_files/abstract.tex
\begin{abstract}
Post-training quantization reduces the memory requirements of vision-language-action (VLA) models, but precision selection must account for the interaction between layer scope, numerical format, and calibration. We introduce \textbf{VLAQuantBench}, a controlled evaluation with 409 runs and 94,574 simulation episodes: four models on LIBERO, with X-VLA additionally evaluated on three simulation benchmark families. Under uncalibrated W4A4 round-to-nearest quantization, expanding a $\pi_{0.5}$ action-head subset from 126 to 167 layers raises success from 7.0\% to 70.5\%. Fixed-observation replay confirms a corresponding numerical recovery. Two-episode calibration removes the severe joint failures in the tested subsets, whereas the same smoothing-and-clipping recipe lowers $\pi_0$ success and does not recover OpenVLA-OFT end-to-end. For OpenVLA-OFT, protecting one 28,672-parameter output projection instead restores near-baseline success: the remaining 441 eligible linear layers retain W3 on LIBERO-Long or eight-bit activations across all four suites. Task-clustered intervals support the large failure and recovery contrasts. These results establish recipe-dependent interactions and identify concrete precision assignments, rather than universal layer-sensitivity rules. Real-kernel and physical-robot measurements complement the accuracy analysis. Code, configurations, and episode records are publicly available.
\end{abstract}

%% file: tex_files/introduction.tex
\section{Introduction}
\label{sec:intro}
Vision-language-action (VLA) models extend pretrained visual and language representations to robot control~\citep{kim2024openvla,black2024pi_0,zheng2025x}. Their memory requirements motivate post-training quantization (PTQ), which lowers numerical precision without retraining the full policy. However, the relevant deployment result is task completion over an execution trajectory. Each action changes the environment and the observations used to choose subsequent actions. A small numerical perturbation can therefore have consequences that are not apparent from an individual forward pass. Therefore, different action sequences can complete the same task.

\input{figs/scope_recovery}

PTQ methods improve low-precision representations through weight reconstruction, activation-aware scaling, and sensitivity-guided precision allocation~\citep{frantar2022gptq,lin2024awq,xiao2023smoothquant,dong2019hawq}. Recent approaches adapt quantization to VLA policies and evaluate their final systems in closed loop~\citep{xu2026qvla,yan2026hbvla,zhang2026quantvla,zheng2026dyq}. These advances raise a complementary evaluation question: which effects follow from the numerical format and the selected layers, and which depend on the quantizer's calibration or optimization? Comparing systems that change checkpoints, scopes, and algorithms together cannot isolate these factors.

A second question concerns the composition of precision decisions. Measuring each layer group separately is an appealing way to screen a large configuration space. Yet an isolated success-rate loss does not necessarily equal that group's marginal effect after other layers have also been quantized. A useful benchmark must therefore include both isolated interventions and their unions. Testing isolated groups alongside their unions directly measures when those sensitivities predict a joint configuration.

We introduce \textbf{VLAQuantBench}, which uses a shared round-to-nearest (RTN) quantizer to compare numerical formats and quantization scopes while retaining each model's released inference conventions. OpenVLA-OFT~\citep{kim2025fine}, $\pi_0$~\citep{black2024pi_0}, $\pi_{0.5}$~\citep{intelligence2025pi_}, and X-VLA~\citep{zheng2025x} are evaluated on LIBERO~\citep{liu2023libero}. X-VLA additionally covers SIMPLER~\citep{li2024evaluating}, CALVIN~\citep{mees2022calvin}, and VLABench~\citep{zhang2025vlabench}. The benchmark comprises 409 evaluation runs and 94,574 simulation episodes, with coverage reported per model and benchmark. VLAQuantBench is also supplemented by layer-group ablations, calibration controls, alternative LLM quantizers, actual kernel measurements, and a real-robot study.

Our main finding is that quantizing additional layers can improve task success and reduce replay error. Format and calibration controls delimit this effect. A complementary intervention makes precision selection constructive: protecting OpenVLA-OFT's output projection recovers failed W3 and A8 configurations.

Our contributions are:
\begin{itemize}[leftmargin=1.3em,itemsep=2pt,topsep=3pt]
\item \textbf{A released evaluation framework.} Model adapters, explicit quantization scopes, benchmark evaluators, and episode records support controlled comparisons and extensions to new policies and quantizers.\footnote{\url{https://github.com/jiuyixu25/VLAQuantBench}}
\item \textbf{Non-monotone and non-additive task sensitivity.} Nested action-head interventions expose recovery and joint failures that isolated tests miss. Held-out replay verifies corresponding numerical effects, while format and calibration controls establish their recipe dependence.
\item \textbf{A targeted remedy for W3 failure.} Keeping one OpenVLA-OFT output projection at released precision restores near-baseline W3 success on Long and A8 success across four suites and local activation statistics alone miss the projection's sensitivity.
\item \textbf{Controls for practical precision selection.} Three-model calibration tests, task-clustered uncertainty, kernel measurements, and physical execution expose choices that change the performance of a precision configuration.
\end{itemize}

%% file: figs/scope_recovery.tex
\begin{figure}[t]
\centering
\includegraphics[width=\linewidth]{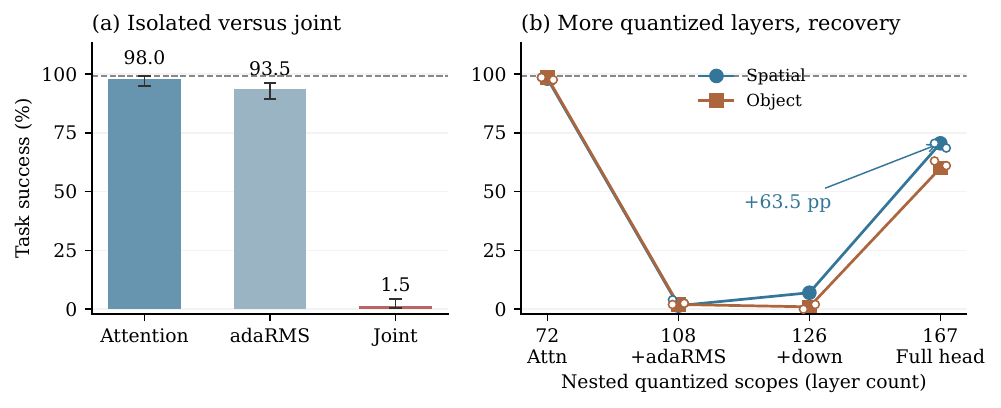}
\caption{Uncalibrated RTN W4A4 action-head interventions in $\pi_{0.5}$. (a) Spatial success for isolated groups and their union; whiskers are 95\% Wilson intervals over 200 episodes. (b) Success along nested scopes. Filled markers show seed 0; open markers show available seed 1/2 repeats. Dashed lines mark the seed-0 baseline (99\%). The 126-to-167-layer expansion improves Spatial success by 63.5 pp, while remaining below baseline. Full subset results are shown in App.~\ref{app:interactions}.}
\label{fig:scope_recovery}
\end{figure}

%% file: tex_files/related_work.tex
\section{Related Work}
\label{sec:related}
\paragraph{Generalist robot policies.}
RT-1 and RT-2 investigate large-scale language-conditioned robot policies and transfer from web representations~\citep{brohan2022rt,zitkovich2023rt}. Open X-Embodiment and Octo further support transfer across robot datasets and embodiments~\citep{o2024open,team2024octo}. OpenVLA, the $\pi$ family, and X-VLA provide different ways to connect pretrained multimodal representations to action generation~\citep{kim2024openvla,black2024pi_0,intelligence2025pi_,zheng2025x}. We evaluate released checkpoints with continuous regression or flow-matching action generation. The present experiments do not include an autoregressive action-token policy.

\paragraph{Post-training quantization.}
Integer inference relies on explicit choices of scale, zero point, and quantization granularity~\citep{jacob2018quantization,krishnamoorthi2018quantizing}. Weight equalization, adaptive rounding, block reconstruction, and stochastic activation quantization improve low-bit approximation~\citep{nagel2019data,nagel2020up,li2021brecq,wei2022qdrop}. GPTQ uses approximate second-order reconstruction for large transformers~\citep{frantar2022gptq}. AWQ uses activation-aware weight scaling~\citep{lin2024awq}. SmoothQuant redistributes channel scales between activations and weights~\citep{xiao2023smoothquant}. LLM.int8() uses mixed-precision treatment of outliers~\citep{dettmers2022gpt3}, while QLoRA introduces NF4 for normally distributed weights in a fine-tuning framework~\citep{dettmers2023qlora}. Our NF4 experiments evaluate the inference representation, not QLoRA training. HAWQ uses sensitivity-guided mixed precision~\citep{dong2019hawq}, while CLADO explicitly models cross-layer dependencies through integer quadratic programming~\citep{deng2025clado}. Cross-layer dependence is therefore already a concern in mixed-precision optimization. Our focus is the non-monotone closed-loop success of nested RTN interventions in robot policies, where the task metric also depends on the execution trajectory.

\paragraph{Efficient VLA models.}
TinyVLA and SmolVLA reduce deployment cost through compact policies~\citep{wen2025tinyvla,shukor2025smolvla}, while BitVLA investigates native low-bit models~\citep{wang2025bitvla}. SQAP-VLA combines quantization and pruning~\citep{fang2025sqap}. VLA-specific PTQ includes channel-sensitive quantization in QVLA~\citep{xu2026qvla}, policy-aware low-bit treatment in HBVLA~\citep{yan2026hbvla}, scale calibration in QuantVLA~\citep{zhang2026quantvla}, and temporal precision adaptation in DyQ-VLA~\citep{zheng2026dyq}. These works already evaluate closed-loop task performance. VLAQuantBench complements them by holding an anchor quantizer fixed and varying numerical format and layer scope. Matched implementations of those methods remain outside the present comparison.

%% file: tex_files/protocol.tex
\section{Benchmark Design}
\label{method}
The benchmark separates three questions: whether a quantized policy completes tasks, which selected layers account for a change in performance, and whether an implementation saves memory or time. Task metrics are primary outcomes, while action deviations and activation statistics are auxiliary diagnostics. Figure~\ref{fig:overview} in App.~\ref{app:quant-prelim} summarizes the evaluation axes.

\paragraph{Models and scopes.}
We partition each policy into a vision encoder (VE), multimodal projector (MP) where present, language backbone (LLM), and action head (AH). Explicit module-root lists and exclusion patterns define disjoint component scopes. Finer scopes select named layer groups or projections. Each result file records the checkpoint, quantization configuration, selected scope, and quantized layer and parameter counts. These records allow numerical results to be traced to the applied intervention. Count checks are supplemented by inspection of activation-hook coverage: a correct weight-layer count alone does not establish that activations were quantized.

\paragraph{Numerical formats.}
$\mathrm{W}b\mathrm{A}a$ denotes $b$-bit linear weights and $a$-bit linear-input activations. Omitting A retains the released activation precision. W2--W4 use asymmetric weight grids with groups of 128 input features per output channel. W8 uses symmetric per-output-channel grids. Short final groups receive their own scale without zero padding. Activations use dynamic symmetric per-token absmax quantization. The six main settings are W3, W4, W8, W4A4, W4A8, and W8A8. 
Targeted studies add W2, A5/A6, and alternative weight configurations. The unquantized baseline is a separate configuration.

RTN accuracy experiments round and dequantize weights in place and quantize linear inputs through forward pre-hooks. Computation remains floating point. Embeddings, biases, normalization, attention matrix products, and the vocabulary head are excluded. X-VLA's embedding-style action decoder is also excluded, while the AH scope covers its quantizable linear layers. These experiments estimate accuracy effects rather than integer-kernel speed or packed-model memory. Actual-kernel measurements are reported separately in \S\ref{sec:efficiency}.

\paragraph{Coverage and native metrics.}
The corpus contains 409 evaluation runs and 94,574 simulation episodes under the selection rule of App.~\ref{app:benchmarks}, including ablations and seed replications. Twenty additional full-precision trajectories are collected for paired-observation replay and reported separately from this closed-loop evaluation count. All four models cover all four evaluated LIBERO suites. X-VLA, whose single LIBERO checkpoint serves every suite, additionally covers the three other benchmark families. LIBERO uses official initial states and the released clients' control conventions, with 200 episodes per configuration or normally 190 for X-VLA. Cross-model baselines retain their own checkpoints and inference settings, so the analysis concerns quantization changes relative to each model's baseline.

SIMPLER visual matching uses 864 episodes per setting. Success is averaged first within each variant, then across variants within a task family, and finally across task families. Available variant-aggregation runs have unequal coverage and are excluded from direct precision comparisons. CALVIN reports completed subtasks per five-task chain over 1,000 chains, reduced to 300 for W4A4. VLABench reports pooled episode progress across five tracks, with 50 episodes per task for baseline/W4/W8, 20 for W3/W4A8/W8A8, and 10 for W4A4; consistently uncompiled scenes are omitted. These metrics are not interchangeable, and we do not combine them into a cross-benchmark average.

\paragraph{Uncertainty and reproducibility.}
Success-rate differences are expressed in percentage points (pp). Reported 95\% Wilson intervals summarize binomial episode-sampling uncertainty, and they do not include uncertainty from training, checkpoint choice, or task clustering. For selected contrasts, we additionally use 5,000 two-stage bootstrap draws, pairing the resampled task identities but resampling episodes independently within each cell and task (App.~\ref{app:replication}). This captures task clustering without preserving episode-level pairing. Interval overlap is not an equivalence test and small point-estimate differences and threshold summaries are descriptive. Selected ablations use additional evaluation seeds and their meaning under fixed initial states is detailed in App.~\ref{app:benchmarks}. We retain task identifiers, episode indices, seeds, outcomes, timing, and checkpoint metadata. The corresponding analysis proves the records source-file hashes and per-cell counts. Protocol checks address simulator version, executed action-chunk length, and activation-hook coverage (\S\ref{sec:pitfalls}).

\paragraph{Public artifacts and extensions.}
The public repository provides evaluation code, model configurations, scope definitions, episode records, and analysis scripts. A new model implements the adapter methods for loading, component selection, task reset, and action prediction. A new quantizer connects to the quantization dispatch while recording its format and affected modules. Existing benchmark evaluators then evaluate the resulting policy against its released-precision baseline. The paper's analyzed subset is specified by its source manifest and repository-wide inventory totals can include other runs.

%% file: tex_files/experiments.tex
\section{Experimental Setup}
\label{sec:experiments}
\paragraph{Policies.}
OpenVLA-OFT combines a LLaMA-2-7B backbone with parallel action decoding and a four-linear-layer L1-regression head~\citep{kim2025fine}. The $\pi_0$ and $\pi_{0.5}$ policies use PaliGemma with flow-matching action experts. $\pi_{0.5}$ additionally uses adaptive normalization conditioning~\citep{black2024pi_0,intelligence2025pi_}. X-VLA uses Florence-2 with a soft-prompted transformer action head~\citep{zheng2025x}. The two $\pi$ models share a backbone family, but their checkpoints and training differ: their comparison is not an intervention that isolates adaptive normalization. The OpenVLA-OFT and X-VLA LIBERO grids run on an RTX 4090 and the two $\pi$ grids run on an RTX 5080 Laptop GPU. The $\pi_0$ results use the LeRobot checkpoint listed in the appendix, not a matched reproduction of every OpenPI-reported policy. Checkpoint identifiers and observed episode budgets are listed in App.~\ref{app:protocol_scale}.

\paragraph{Experiment groups.}
The end-to-end grid quantizes all eligible linear layers at the six main settings. Component experiments quantize VE, LLM, or AH alone, extending weights to W2 and activations to A5/A6 where available. Action-head ablations compare individual projection groups, unions, and complements. The central $\pi_{0.5}$ analysis includes a 14-subset table and additional attention/modulation combinations. These are controlled interventions within a fixed checkpoint and task suite. The six-setting $\pi_{0.5}$ comparison is repeated at W4 and W4A8; a separate OpenVLA-OFT-Long W3 study isolates and protects its output projection. Same-observation replay covers 20 held-out $\pi_{0.5}$ trajectories across all ten Spatial tasks, with matched flow noise and three numerical formats.

\paragraph{Alternative quantizers and calibration.}
AWQ, NF4, LLM.int8(), and SmoothQuant are evaluated on OpenVLA-OFT's LLM with matched-bit RTN references. Implementations retain their method-specific calibration and numerical formats; equal nominal bit width does not make their representations identical. A separate SmoothQuant-style experiment uses two unquantized robot episodes to study smoothing and clipping under different quantized scopes. It is a calibration control, not an optimized method benchmark. The LLM SmoothQuant deployment path uses $\alpha=0.85$ and 512 text calibration samples; the scope-control experiment instead uses $\alpha=0.5$ or $0.75$ and robot observations. 

\paragraph{Deployment and physical execution.}
Real-kernel profiling uses OpenVLA-OFT on one RTX 4090, batch size one, and 20 LIBERO-Spatial rollouts per method. We report both control-call and complete-network-inference timings, with the aggregation defined in \S\ref{sec:efficiency}. Separately, OpenVLA-OFT and $\pi_{0.5}$ are LoRA-finetuned~\citep{hu2022lora} using 200 demonstrations each and evaluated on Pick and Pick-and-place with a Franka Research~3. The physical study contains BF16 and five quantized settings, with 50 rollouts per model--task--precision condition. These policies and tasks differ from the simulation checkpoints and suites.

%% file: tex_files/discussion.tex
\section{Results}
\label{sec:results}
\input{tabs/exp1_libero}
\input{tabs/exp1_xvla_cross}
\subsection{Eight-bit activations can cause severe task losses}
\label{sec:e2e}
Weight-only W8 closely tracks the LIBERO baselines in Table~\ref{tab:exp1_libero}, whereas W4 is not uniformly lossless. OpenVLA-OFT's Goal success falls from 96.0\% to 82.0\%. The W3 result on OpenVLA-OFT-Long is particularly poor (8.0\%, compared with 88.5\% at W4). Thus, conclusions drawn from an average across suites can hide substantial losses in individual settings.

Activation quantization produces larger differences. OpenVLA-OFT's W4A8 success is 21.5/2.5/2.0/76.0\% on Spatial/Object/Goal/Long. Both tasks and checkpoints vary across these comparisons, so they do not isolate a checkpoint effect from task difficulty. Under W4A4, $\pi_0$ retains 31--74\% success, whereas the other three models reach zero on all four suites. Absolute success and the change from each model's baseline provide complementary comparisons. X-VLA uses the same LIBERO checkpoint across all four suites: every weight-only and A8 format stays within 3.2 pp of its baseline on each suite, and W4A4 is zero on each (Table~\ref{tab:exp1_libero}), showing that W4A4 failure recurs across the four evaluated suites with the checkpoint held fixed.

For X-VLA, W4 and W4A8 remain close to their baselines on SIMPLER, CALVIN, and VLABench, while W4A4 sharply degrades all three (Table~\ref{tab:exp1_xvla}). This model-specific extension tests whether the pattern persists beyond LIBERO.

\input{tabs/components}
\paragraph{Component sensitivity.}
\label{sec:components}
Component-only W4A4 yields different sensitivity orderings (Table~\ref{tab:components}). For $\pi_{0.5}$, LLM-only quantization gives zero success while AH-only retains 70.5\%. X-VLA reverses this ordering: 96.3\% with LLM-only and zero with AH-only. OpenVLA-OFT reaches zero under each component intervention. Protecting the same component across models therefore misses the observed differences. Descriptive precision thresholds are provided in App.~\ref{app:screening}.

\subsection{Quantizing more layers can improve success}
\label{sec:lattice}
Figure~\ref{fig:scope_recovery} compares nested uncalibrated W4A4 RTN interventions within $\pi_{0.5}$'s action head. The 126-layer set containing attention, adaRMS, and down projections scores 7.0\% on Spatial; its 167-layer full-head superset scores 70.5\%. Quantizing 41 additional layers raises success by 63.5 pp, although it remains 28.5 pp below the unquantized baseline. The added layers comprise gate/up projections and five other linear layers. Full-head success repeats at 70.5/70.5/68.5\%. On Object, the 126-layer set scores 1.0/0.0/2.0\% across three seeds and the full head 60.0/63.0/61.0\%, giving 59--63 pp of recovery (App.~\ref{app:interactions}). Thus, success need not decrease as the quantized scope expands.

\paragraph{Isolated losses also underestimate joint failure.}
Let $S(Q)$ denote success when layer set $Q$ is quantized, and $D(Q)=S(\varnothing)-S(Q)$. For disjoint groups, an additive success-loss approximation predicts $D(A\cup B)\approx D(A)+D(B)$. Spatial attention and adaRMS interventions lose 1.0 and 5.5 pp individually, but their union loses 97.5 pp. Union success remains low across seeds (1.5/4.0/2.5\%) and on Object (2.0/2.0/2.5\%). X-VLA's two MLP projection groups lose 30.5 and 2.6 pp separately and 72.6 pp jointly. Tables~\ref{tab:lattice}--\ref{tab:interaction_controls} provide the complete subset and supporting comparisons.

\paragraph{The reversal depends on the quantization recipe.}
The same six configurations at W4 and W4A8 score 97.5--100.0\%, all within 1.5 pp of the 99.0\% baseline (Table~\ref{tab:subsets_formats}). In particular, the union and 126-layer set no longer collapse. The strong effects are therefore not universal properties of W4 weights: they depend on activation precision in this checkpoint and scope family. Calibration further changes this behavior: smoothing and clipping from two held-out episodes raises union success from 1.5\% to 97.5\% and the 126-layer subset from 7.0\% to 99.0\% (Table~\ref{tab:calibrated_subsets}). The severe joint failures therefore belong to the tested uncalibrated per-token absmax A4 recipe, not to all W4A4 policies.

\input{figs/replay_formats}
\paragraph{Numerical interaction before environment feedback.}
Success-rate non-additivity alone could reflect a task-failure threshold. We therefore replay 20 held-out full-precision trajectories with identical observations and flow noise for every configuration, giving 219 predicted chunks. Repeated full-precision replay exactly reproduces the recorded actions. Let $e_Q$ be the flattened normalized action-chunk difference from full precision after the complete ten-step flow solve. At W4A4, the median ratio $\|e_{A\cup B}\|_2/(\|e_A\|_2+\|e_B\|_2)$ is 2.90 (interquartile range 2.27--3.50), and the residual $\|e_{A\cup B}-e_A-e_B\|_2/\|e_{A\cup B}\|_2$ is 0.82. These output errors cannot be explained by simple vector addition of the isolated perturbations.

Recovery also precedes environment feedback: expanding 126 to 167 layers lowers normalized chunk MAE from 0.514 to 0.390, with lower error on 217 of 219 chunks (99.1\%) and lower mean error in each of the 20 trajectories (Figure~\ref{fig:replay_formats}). At W4/W4A8, the corresponding MAE instead increases from 0.014/0.015 to 0.023/0.024, with no chunk improving. Their union errors are much closer to vector addition (relative residuals 0.17/0.20; cosine similarities 0.99/0.98). Thus, the tested A4 reversal appears in both numerical outputs and closed-loop outcomes, rather than only in the thresholded task metric.

\paragraph{Five additional layers recover more than the gate/up group.}
Splitting the 41 added layers reveals where recovery occurs (Table~\ref{tab:calibrated_subsets}). Adding the 36 gate/up projections to the 126-layer subset yields 26.5\% success, while adding only the five remaining conditioning/input/output projections yields 78.5\%. The latter adds 5.3M parameters and reduces executed-action MAE from 0.138 to 0.089 in the same-observation replay. Gate/up quantization also improves success, by 19.5 pp, but does not produce the larger recovery. The 131-layer result is not established as better than the full head: their 8-pp difference has a task-clustered interval spanning zero. These interventions identify an effective small group without proving a particular compensation mechanism.
\input{tabs/calibrated_subsets_pi05}

\paragraph{Diagnostics narrow the explanations.}
For W4A4, signed gripper-command bias is $+0.073$ for the union, $-0.057$ for 126 layers, and $-0.004$ for the full head. The down-projection input per-token absmax 99th-percentile summaries remain close for 126 and 167 layers (14.30 and 14.18). This aggregate statistic does not track the large recovery; it does not exclude layer- or timestep-specific outlier effects. Denoising-state deviations generally build during the flow solve, but are not strictly monotone. App.~\ref{app:pi05_replay} gives the measurement definitions and full diagnostics. The replay establishes output-level interaction; the causal roles of directional bias, activation ranges, and individual flow steps remain to be separated.

\input{tabs/culprit}
\subsection{Small projections and local proxies hide task sensitivity}
\label{sec:culprit}
On OpenVLA-OFT-Spatial, W4A8 on the final AH projection, \texttt{fc2}, gives 18.0\% success (95\% Wilson interval 13--24\%), compared with 20.5\% (15--27\%) for all four AH layers and 100.0\% at baseline. This projection has 28,672 parameters, about 0.019\% of the head. Quantizing only the first projection or only the two residual layers retains 99.0\% and 99.5\%, respectively. The \texttt{fc2}-only result repeats at 17.0\%. Quantizing this small output projection alone is sufficient to cause an 82-pp loss, comparable to quantizing the entire head.

\paragraph{Protecting one layer recovers end-to-end W3 performance.}
On OpenVLA-OFT-Long, end-to-end W3 gives 8.0\% success against a 92.5\% baseline. Quantizing only the final action projection gives 12.5\% and 10.5\% across two evaluation seeds, while quantizing all vision, projector, and language layers retains 95.0\% (Table~\ref{tab:oft_long_w3}). Keeping only that projection at released precision among the eligible linear layers yields 93.5\% [89,96], with the remaining 441 of 442 layers quantized at W3. Two further evaluation seeds give 91.5\% and 91.0\% for this protected-output assignment. The recorded counts are VE 209, MP 5, LLM 224, and AH 3; embeddings and other excluded operations retain their usual precision. This is an accuracy result using simulated quantization, not a packed W3 deployment measurement.

Reducing the action head's weight-group size from 128 to 64 or 32 gives only 12.5\% and 17.5\%. These tested grid refinements do not recover near-baseline success, whereas protecting the output projection does. The result supplies a concrete precision-allocation remedy without ruling out gains from other quantizers. The sensitive projection has the same architectural role as in the Spatial W4A8 experiment, but the checkpoints and tasks differ, so the contrast does not isolate a horizon effect.
\input{tabs/oft_long_w3_localization}

The same protection rule also resolves the severe OpenVLA-OFT A8 failures across all four LIBERO suites (Table~\ref{tab:oft_protect_fc2}). Protected W4A8 scores 97.5/97.0/98.0/94.0\% on Spatial/Object/Goal/Long, and protected W8A8 scores 99.0/97.5/97.5/93.5\%. All are within 2.5 pp of their corresponding baselines. This supports protecting that output projection in these tested checkpoints and formats; it does not establish a universal rule for other action heads.
\input{tabs/oft_protect_fc2_a8}

For $\pi_{0.5}$, quantizing its action-output projection at W4A4 retains 97.0\% against a 99.0\% baseline. To test whether multiple integration steps are necessary for this tolerance, we also reduce the flow-matching step count. With one step, the quantized projection gives 98.0\% against its matched 98.5\% baseline; with two steps, 99.0\% against 97.0\%. The projection therefore remains tolerant even with a single flow-matching step.

X-VLA differs again: quantizing its qkv group gives 11.6\% while the output-projection group gives 96.3\%. The groups contain 75.5M and 25.2M parameters, respectively. Across models, neither parameter count nor architectural role supplies a universal precision rule.

\paragraph{Nearly identical kurtosis, sharply different outcomes.}
\label{sec:proxy}
OpenVLA-OFT provides a direct counterexample to ranking these groups by activation kurtosis alone. Its final projection and residual-layer group have kurtosis 834.3 and 835.3, but W4A8 success is 18.0\% and 99.5\%, respectively: an 81.5-pp gap under the same checkpoint, suite, and numerical format. This comparison shows how a nearly identical local statistic can accompany very different closed-loop sensitivity.

OpenVLA-OFT's residual blocks add their transformed features to a skip path, whereas its final projection directly produces action coordinates. This suggests that error placement, not just the input distribution, matters. The architecture does not force the two measured kurtoses to be equal, and the skip-path explanation remains a hypothesis. The familiar practice of preserving output layers fits OpenVLA-OFT, while $\pi_{0.5}$'s tolerant output projection shows why the rule still needs model-specific testing. Small-sample rank summaries are retained as diagnostics in App.~\ref{app:diagnostics}.

\input{tabs/c2_granularity}
\subsection{Calibration is part of the configuration}
\label{sec:c2}
On $\pi_{0.5}$-Spatial, end-to-end W3 yields 0.0\% with symmetric per-channel weights and 20.0\% with asymmetric per-channel weights. Group-64 grids instead yield 97.5\% and 97.0\%, respectively. Asymmetry alone therefore does not restore near-baseline performance in the per-channel setting. W4 gives 97--98\% across the tested grids (Table~\ref{tab:c2}). The asymmetric group-128 entries reuse the main-grid results, with the same checkpoint, scope, and quantization preset.

\input{tabs/calibration_three_models}
\paragraph{The recipe does not transfer uniformly across models.}
Two-episode SmoothQuant-style smoothing and 99.9th-percentile clipping have different effects across models and scopes (Table~\ref{tab:calibration_three_models}). For $\pi_{0.5}$, VE-only W4A4 improves from 37.0\% to 80.0\%, and the tested action-head unions recover near-baseline success. For $\pi_0$, the recipe lowers every measured scope: LLM-only drops from 72.0\% to 11.0\%, AH-only from 58.0\% to 39.5\%, and end-to-end from 61.0\% to 10.5\% at $\alpha=0.5$. End-to-end results at $\alpha=0.25/0.75$ are 9.0/8.0\% (Table~\ref{tab:pi0_calibration}). OpenVLA-OFT recovers partially in LLM-only and AH-only scopes (0.0\% to 33.0/38.5\%), but VE-only remains zero; end-to-end scores are 0.0/9.5\% at $\alpha=0.5/0.75$.

For $\pi_{0.5}$, raising end-to-end smoothing strength to $\alpha=0.75$ yields 91.5\%, whereas $\alpha=0.5$ yields zero. The original AH/full-policy calibrations used states 0--1, overlapping evaluation; replications with three disjoint two-episode sets yield 97.0\% AH-only and 88.0--90.5\% end-to-end (Table~\ref{tab:c3_stability}). New component and subset controls use states 20--21. The table distinguishes reused historical cells from new controls and records model-specific calibration coverage. These are controlled recipe tests, not optimized method comparisons. Calibration must be validated together with the complete precision assignment.

\paragraph{Uncertainty in the large contrasts.}
Task-clustered bootstrap intervals preserve the large uncalibrated failure and recovery effects: Spatial union versus baseline is $-97.5$ pp with a 95\% interval of $[-100.0,-93.0]$, and full-head versus 126-layer success is $+63.5$ pp with $[46.0,80.5]$ (Table~\ref{tab:clustered}). Small differences near baseline are less resolved. These intervals resample the ten observed tasks; they are not evidence of generalization to arbitrary tasks or equivalence to the baseline.

\subsection{Near-ceiling tasks leave action-fidelity differences unresolved}
\label{sec:methods_res}
Alternative LLM quantizers on OpenVLA-OFT reach 98.0--99.0\% success, close to matched-bit RTN (Table~\ref{tab:methods}). On one held-out trajectory, AWQ reduces mean absolute action deviation relative to RTN by factors of 1.44 at W4 and 1.15 at W3 (Table~\ref{tab:fidelity}). This comparison reveals action differences that the near-ceiling task scores cannot resolve; it cannot test whether lower action error predicts higher success in failure-prone settings. App.~\ref{app:fidelity} reports deviations per action coordinate. Unlike this near-ceiling LLM comparison, the $\pi_{0.5}$ replay in \S\ref{sec:lattice} includes severe-failure configurations and tests numerical composition directly.

\input{tabs/efficiency}
\subsection{Memory reduction does not imply lower latency}
\label{sec:efficiency}
Table~\ref{tab:efficiency} reports real LLM kernels on an RTX 4090 at batch size one. All tested quantized paths reduce peak allocated memory and take longer than BF16 under both control-call and full-inference timing. Control calls can serve cached actions; full inference generates a new chunk. Timing aggregation, warm-up, synchronization, and memory units are defined in App.~\ref{app:kernel}.

Three independent repeat sessions preserve the finding that all quantized paths are slower than BF16. Across those repeats, the largest within-method ranges are 0.38\,ms for control calls and 3.00\,ms for full inference (Table~\ref{tab:efficiency_sessions}). AWQ and SmoothQuant have close control-call times and exchange order across sessions; a stable ordering between them is not established. OpenVLA-OFT decodes an action chunk in a parallel forward pass, rather than through single-token autoregressive decoding. The balance between matrix computation, online quantization, and weight dequantization can therefore differ from decoding workloads where weight-only kernels save time; kernel-level profiling would be needed to attribute the measured overhead.

\input{tabs/exp4}
\subsection{Physical-robot precision comparisons}
\label{sec:real}
Table~\ref{tab:exp4} reports the physical-robot precision comparisons. W3 substantially reduces success for both models and tasks, while W8 and W8A8 remain within 6 pp of the corresponding BF16 baselines. With W4 weights, adding A8 reduces success by 18--40 pp across the four model--task settings.

\FloatBarrier
\section{Evaluation-Stack Checks}
\label{sec:pitfalls}
Three implementation checks are essential to reproducing the evaluated policy.
\paragraph{Simulator version.}
Changing MuJoCo versions altered the settling of a stored LIBERO initial state. We pin MuJoCo to 3.3.2 for OpenVLA-OFT and the $\pi$ models and examine per-task baselines; the X-VLA environment pins MuJoCo 3.1.6, which predates the change, and the version is recorded with its diagnostics. The version and per-task baseline are recorded with the evaluation configuration.
\paragraph{Executed action chunks.}
Executing 50 actions before re-observing instead of the intended 10 changed the tested $\pi$-family control loop and reduced success. Predicted chunk length and executed action count must be distinguished. Other clients retain their own execution conventions; OpenVLA-OFT's simulation adapter, for example, uses an eight-action chunk.
\paragraph{Activation coverage.}
An activation-quantization run can silently become weight-only for a component if its input hooks are omitted. Missing AH hooks made OpenVLA-OFT appear much more tolerant in earlier internal runs. Selected module names, layer counts, and executed hooks must therefore be checked together. Together, these checks make the executed quantization and control settings traceable.

%% file: tabs/exp1_libero.tex
\begin{table*}[htbp]
  \centering\small
  \setlength{\tabcolsep}{5.5pt}
  \begin{tabular}{llccccccc}
    \toprule
    \textbf{Model} & \textbf{Suite} & \textbf{Baseline} & \textbf{W3} & \textbf{W4} & \textbf{W8} & \textbf{W4A4} & \textbf{W4A8} & \textbf{W8A8} \\
    \midrule
    \multirow{4}{*}{$\pi_{0.5}$} & Spatial & 99.0 & 99.0 & 97.5 & 98.5 & \textbf{0.0} & 99.0 & 98.5 \\
     & Object & 99.0 & 100.0 & 99.0 & 99.0 & \textbf{0.0} & 100.0 & 99.0 \\
     & Goal & 94.0 & 95.5 & 96.5 & 96.0 & \textbf{0.0} & 96.5 & 97.0 \\
     & Long (10) & 95.0 & 94.5 & 94.0 & 96.5 & \textbf{0.0} & 96.5 & 93.5 \\
    \midrule
    \multirow{4}{*}{$\pi_0$} & Spatial & 78.0 & 71.0 & 69.0 & 77.5 & 61.0 & 64.5 & 74.5 \\
     & Object & 87.5 & 76.0 & 90.5 & 86.5 & 74.0 & 88.5 & 88.5 \\
     & Goal & 84.5 & 64.5 & 79.0 & 83.0 & 58.0 & 79.0 & 81.0 \\
     & Long (10) & \textbf{48.5} & \textbf{40.0} & \textbf{41.0} & \textbf{47.5} & \textbf{31.0} & \textbf{41.0} & \textbf{45.0} \\
    \midrule
    \multirow{4}{*}{OpenVLA-OFT} & Spatial & 100.0 & 78.0 & 96.0 & 99.5 & \textbf{0.0} & \textbf{21.5} & \textbf{36.0} \\
     & Object & 97.5 & 74.5 & 93.5 & 98.0 & \textbf{0.0} & \textbf{2.5} & \textbf{2.5} \\
     & Goal & 96.0 & 71.5 & 82.0 & 97.5 & \textbf{0.0} & \textbf{2.0} & \textbf{13.0} \\
     & Long (10) & 92.5 & \textbf{8.0} & 88.5 & 93.5 & \textbf{0.0} & 76.0 & 64.0 \\
    \midrule
    \multirow{4}{*}{X-VLA} & Spatial & 96.8 & 96.8 & 95.8 & 97.4 & \textbf{0.0} & 96.3 & 96.3 \\
     & Object & 97.9 & 96.8 & 97.9 & 97.9 & \textbf{0.0} & 98.4 & 97.9 \\
     & Goal & 97.4 & 97.4 & 96.8 & 96.3 & \textbf{0.0} & 97.9 & 96.3 \\
     & Long (10) & 97.9 & 94.7 & 95.8 & 97.4 & \textbf{0.0} & 96.8 & 98.4 \\
    \bottomrule
  \end{tabular}
  \caption{End-to-end RTN on LIBERO (success \%). Each setting uses 200 episodes, except X-VLA with 190. All formats use the grids in Section~\ref{method}. Bold entries are below 50\%; this visual threshold is not a significance test.}
  \label{tab:exp1_libero}
\end{table*}

%% file: tabs/exp1_xvla_cross.tex
\begin{table*}[htbp]
  \centering\small
  \begin{tabular}{lccccccc}
    \toprule
    \textbf{Benchmark (metric)} & \textbf{Baseline} & \textbf{W3} & \textbf{W4} & \textbf{W8} & \textbf{W4A4} & \textbf{W4A8} & \textbf{W8A8} \\
    \midrule
    SIMPLER-VM (SR \%) & 67.6 & 64.6 & 65.6 & 67.5 & 3.9 & 67.7 & 67.1 \\
    CALVIN ABC$\rightarrow$D (Avg.\ Len.) & 4.28 & 4.13 & 4.20 & 4.25 & 0.00 & 4.20 & 4.27 \\
    VLABench (Avg.\ PS \%) & 35.0 & 33.1 & 32.7 & 34.8 & 0.0 & 34.0 & 33.6 \\
    \bottomrule
  \end{tabular}
  \caption{X-VLA beyond LIBERO. SIMPLER-VM averages variants within task families and then averages families (864 episodes per setting). CALVIN reports completed subtasks out of five (1,000 chains, or 300 for W4A4). VLABench reports mean progress in percent with the budgets in App.~\ref{app:protocol_scale}. Unequal-coverage SIMPLER-VA runs are excluded.}
  \label{tab:exp1_xvla}
\end{table*}

%% file: tabs/components.tex
\begin{table}[htbp]
\centering\small
\begin{tabular}{lrrrr}
\toprule
Model & Baseline & VE & LLM & AH \\
\midrule
$\pi_{0.5}$ & 99.0 & 37.0 & 0.0 & 70.5 \\
$\pi_0$ & 78.0 & 76.5 & 72.0 & 58.0 \\
OpenVLA-OFT & 100.0 & 0.0 & 0.0 & 0.0 \\
X-VLA & 96.8 & 67.9 & 96.3 & 0.0 \\
\bottomrule
\end{tabular}
\caption{Component-only W4A4 on LIBERO-Spatial (success \%). Only the indicated component is quantized; baseline denotes the released-precision policy.}
\label{tab:components}
\end{table}

%% file: figs/replay_formats.tex
\begin{figure}[t]
\centering
\includegraphics[width=\linewidth]{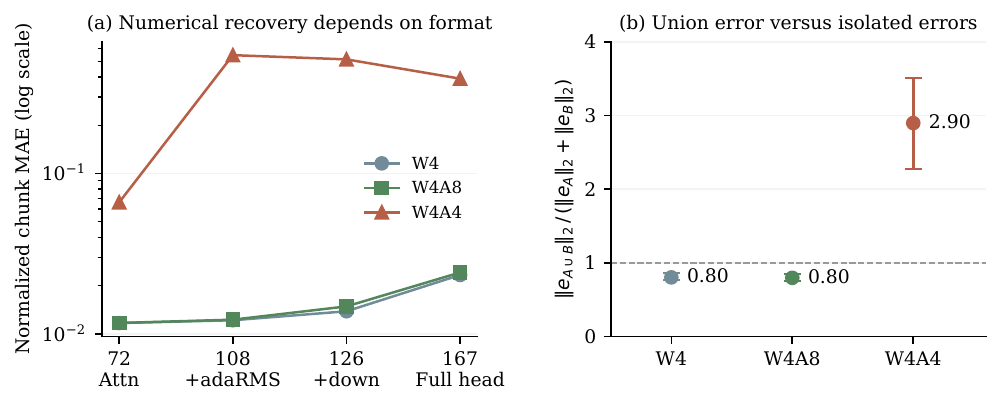}
\caption{Uncalibrated RTN same-observation, same-noise replay on 20 held-out $\pi_{0.5}$ trajectories (219 predicted chunks). (a) W4A4 recovery is present in normalized action-chunk error before environment feedback; W4/W4A8 do not show the 126-to-167-layer reversal. (b) Points and whiskers give the median and interquartile range of the per-chunk error-norm ratio, not confidence intervals. $A$ denotes attention and $B$ adaRMS. The dashed value 1 is an upper bound under exact vector addition; cancellation can give smaller ratios.}
\label{fig:replay_formats}
\end{figure}

%% file: tabs/calibrated_subsets_pi05.tex
\begin{table}[t]
\centering
\small
\setlength{\tabcolsep}{3pt}
\begin{tabular}{lrrrrr}
\toprule
AH subset & \#L & RTN & Calibrated & Exec. MAE & Gripper bias \\
\midrule
full precision & 0 & 99.0 [96,100] & -- & -- & -- \\
attention & 72 & 98.0 [95,99] & 98.5 [96,99] & 0.012 & +0.005 \\
adaRMS & 36 & 93.5 [89,96] & 97.5 [94,99] & 0.037 & +0.007 \\
attention + adaRMS & 108 & 1.5 [1,4] & 97.5 [94,99] & 0.125 & +0.073 \\
+ down\_proj & 126 & 7.0 [4,11] & 99.0 [96,100] & 0.138 & -0.057 \\
126 + gate/up & 162 & 26.5 [21,33] & -- & 0.129 & +0.051 \\
126 + five other layers & 131 & 78.5 [72,84] & -- & 0.089 & -0.019 \\
full action head & 167 & 70.5 [64,76] & 98.5 [96,99] & 0.089 & -0.004 \\
\bottomrule
\end{tabular}
\caption{$\pi_{0.5}$-Spatial W4A4 scope controls (SR \%, 200 episodes; Wilson intervals). New isolated/union/126-layer calibrations use task-0 states 20--21; the full-head calibrated cell is reused from the original states-0--1 experiment. Smoothing and clipping act on selected AH layers at $\alpha=0.5$. The two split rows add either 36 gate/up projections or five conditioning/input/output projections to the same 126-layer subset. Replay columns describe the uncalibrated configurations only; dashes denote unmeasured entries. The 131-layer split differs from the 131-layer ``all but adaRMS scope in Table~
ef{tab:lattice}.}
\label{tab:calibrated_subsets}
\end{table}

%% file: tabs/culprit.tex
\begin{table}[htbp]
  \centering\small
  \setlength{\tabcolsep}{5pt}
  \begin{tabular}{lrrr}
    \toprule
    \textbf{Quantized layers} & \textbf{Params} & \textbf{SR \%} & $\Delta$ \\
    \midrule
    \multicolumn{4}{l}{\emph{OpenVLA-OFT, W4A8, baseline 100.0\%}} \\
    \quad \texttt{fc1} & 117.4M & 99.0 & -1.0 \\
    \quad residual blocks (2) & 33.6M & 99.5 & -0.5 \\
    \quad \textbf{\texttt{fc2}} & \textbf{28.7K} & 18.0 & -82.0 \\
    \quad all four layers & 151.0M & 20.5 & -79.5 \\
    \midrule
    \multicolumn{4}{l}{\emph{X-VLA, W4A4, baseline 96.8\%}} \\
    \quad \textbf{\texttt{attn.qkv}} (24) & 75.5M & 11.6 & -85.3 \\
    \quad \texttt{attn.proj} (24) & 25.2M & 96.3 & -0.5 \\
    \quad \texttt{mlp.fc1} (24) & 100.7M & 66.3 & -30.5 \\
    \quad \texttt{mlp.fc2} (24) & 100.7M & 94.2 & -2.6 \\
    \quad all 98 layers & 304.1M & 0.0 & -96.8 \\
    \midrule
    \multicolumn{4}{l}{\emph{$\pi_{0.5}$, W4A4, baseline 99.0\%}} \\
    \quad \texttt{action\_out\_proj} & 32.8K & 97.0 & -2.0 \\
    \bottomrule
  \end{tabular}
  \caption{Layer-group interventions on LIBERO-Spatial. Parameters count selected linear weights; $\Delta$ is the success change from the corresponding baseline in pp. OFT and $\pi_{0.5}$ use 200 episodes; X-VLA uses 190, except its MLP fc2 group with 191. Differences are computed from unrounded success rates before rounding to one decimal place.}
  \label{tab:culprit}
\end{table}

%% file: tabs/oft_long_w3_localization.tex
\begin{table}[t]
\centering
\small
\begin{tabular}{llrrr}
\toprule
Quantized scope & Grid & \#L & Params & SR \% [CI] \\
\midrule
none (baseline) & -- & 0 & -- & 92.5 [88,95] \\
end-to-end & W3 & 442 & 7.44B & 8.0 [5,13] \\
LLM only & W3 & 224 & 6.48B & 92.5 [88,95] \\
VE only & W3 & 209 & 728M & 95.5 [92,98] \\
VE + MP + LLM (all but the AH) & W3 & 438 & 7.29B & 95.0 [91,97] \\
AH only & W3 & 4 & 151.0M & 9.5 [6,14] \\
\quad fc1 only & W3 & 1 & 117.4M & 94.0 [90,97] \\
\quad residual blocks only & W3 & 2 & 33.6M & 94.0 [90,97] \\
\quad fc2 only & W3 & 1 & 28.7K & 12.5 [9,18] \\
\quad fc2 only, evaluation seed 1 & W3 & 1 & 28.7K & 10.5 [7,16] \\
AH only & W3, group 64 & 4 & 151.0M & 12.5 [9,18] \\
AH only & W3, group 32 & 4 & 151.0M & 17.5 [13,23] \\
\midrule
all but AH fc2 (441/442 layers) & W3 & 441 & 7.44B & 93.5 [89,96] \\
\quad evaluation seed 1 & W3 & 441 & 7.44B & 91.5 [87,95] \\
\quad evaluation seed 2 & W3 & 441 & 7.44B & 91.0 [86,94] \\
\bottomrule
\end{tabular}
\caption{OFT-Long W3 localization (success \%, 200 episodes per cell; 95\% Wilson intervals). W3 uses asymmetric group-128 weights unless specified. The final three rows protect only the action head's output projection among eligible linear layers: 441 layers remain quantized. Other non-linear/excluded operations retain released precision. Layer counts are checked against run metadata.}
\label{tab:oft_long_w3}
\end{table}

%% file: tabs/oft_protect_fc2_a8.tex
\begin{table}[t]
\centering
\small
\setlength{\tabcolsep}{4pt}
\begin{tabular}{lrrrrr}
\toprule
& & \multicolumn{2}{c}{W4A8} & \multicolumn{2}{c}{W8A8} \\
Suite & Baseline & All layers & Protect fc2 & All layers & Protect fc2 \\
\midrule
Spatial & 100.0 & 21.5 & 97.5 [94,99] & 36.0 & 99.0 [96,100] \\
Object & 97.5 & 2.5 & 97.0 [94,99] & 2.5 & 97.5 [94,99] \\
Goal & 96.0 & 2.0 & 98.0 [95,99] & 13.0 & 97.5 [94,99] \\
Long & 92.5 & 76.0 & 94.0 [90,97] & 64.0 & 93.5 [89,96] \\
\bottomrule
\end{tabular}
\caption{OFT protected-output comparisons (SR \%, 200 episodes; Wilson intervals for protected cells). Both weights and input activations of the AH output projection remain at released precision; 441 other eligible linear layers are quantized (VE 209, MP 5, LLM 224, AH 3). Protected scores are within 2.5 pp of the respective baselines across the tested formats and suites. The W3 Long assignment yields 93.5/91.5/91.0\% across three seeds.}
\label{tab:oft_protect_fc2}
\end{table}

%% file: tabs/c2_granularity.tex
\begin{table}[htbp]
  \centering\small
  \begin{tabular}{llcc}
    \toprule
    \textbf{Bits} & \textbf{Granularity} & \textbf{Sym.} & \textbf{Asym.} \\
    \midrule
    \multirow{3}{*}{W3} & per-channel & 0.0 & 20.0 \\
     & group-64 & 97.5 & 97.0 \\
     & group-128 & 96.0 & 99.0 \\
    \midrule
    \multirow{3}{*}{W4} & per-channel & 98.0 & 97.5 \\
     & group-64 & 97.0 & 97.5 \\
     & group-128 & 98.0 & 97.5 \\
    \bottomrule
  \end{tabular}
  \caption{Weight-grid controls for end-to-end $\pi_{0.5}$ on LIBERO-Spatial (success \%, 200 episodes). Sym./Asym. denote symmetric/asymmetric grids. Asymmetric group-128 entries are the main presets from Table~\ref{tab:exp1_libero}, reused here rather than newly measured. W4 point estimates span 97--98\%.}
  \label{tab:c2}
\end{table}

%% file: tabs/calibration_three_models.tex
\begin{table}[t]
\centering
\small
\setlength{\tabcolsep}{4pt}
\begin{tabular}{lrrrrrr}
\toprule
& \multicolumn{2}{c}{$\pi_{0.5}$ (99.0)} & \multicolumn{2}{c}{$\pi_0$ (78.0)} & \multicolumn{2}{c}{OFT (100.0)} \\
Quantized scope & RTN & calibrated & RTN & calibrated & RTN & calibrated \\
\midrule
VE only & 37.0 & 80.0 & 76.5 & 72.0 & 0.0 & 0.0 \\
LLM only & 0.0 & 70.5 & 72.0 & 11.0 & 0.0 & 33.0 \\
AH only & 70.5 & 98.5 & 58.0 & 39.5 & 0.0 & 38.5 \\
end-to-end, $\alpha$=0.5 & 0.0 & 0.0 & 61.0 & 10.5 & 0.0 & 0.0 \\
end-to-end, $\alpha$=0.75 & 0.0 & 91.5 & 61.0 & 8.0 & 0.0 & 9.5 \\
\bottomrule
\end{tabular}
\caption{W4A4 smoothing-and-clipping controls on LIBERO-Spatial (SR \%, 200 episodes). Component-only settings use $\alpha=0.5$; the final two rows specify end-to-end strength. New cells use task-0 states 20--21. The $\pi_{0.5}$ LLM/AH/end-to-end calibrated values reuse the original states-0--1 cells; its VE result (80.0\%) is new. MP is not calibrated for the $\pi$ models; OFT end-to-end includes MP and observes 204/209 VE layers. Coverage and disjoint-set controls are detailed in App.~\ref{app:calib}.}
\label{tab:calibration_three_models}
\end{table}

%% file: tabs/efficiency.tex
\begin{table}[htbp]
\centering\small
\begin{tabular}{lrrr}
\toprule
Method & Memory (GiB) & Control call (ms) & Full inference (ms) \\
\midrule
BF16 & 15.0 & 18.8 & 139.9 \\
AWQ & 6.1 & 24.1 & 181.1 \\
NF4 & 6.3 & 20.2 & 152.6 \\
LLM.int8() & 9.0 & 48.3 & 379.5 \\
SmoothQuant & 9.0 & 23.8 & 181.8 \\
\bottomrule
\end{tabular}
\caption{Real-kernel measurements on OpenVLA-OFT (LLM only, RTX 4090, batch size one, 20 LIBERO episodes). Memory is peak CUDA allocation. Timings are medians of episode means. Control calls include cached actions; full inference generates a new action chunk. See App.~\ref{app:kernel} for timer definitions.}
\label{tab:efficiency}
\end{table}

%% file: tabs/exp4.tex
\begin{table}[htbp]
\setlength{\tabcolsep}{3pt}
\renewcommand{\arraystretch}{1.}
  \centering
  \small
    \begin{tabular}{cccccccc}
    \toprule
    \multirow{2}{*}{\textbf{Task}} & \multirow{2}{*}{\textbf{Model}} & \multicolumn{6}{c}{\textbf{SR (\%)}} \\
    \cmidrule(lr){3-8}
     & & BF16 & W3 & W4 & W8 & W4A8 & W8A8 \\
    \midrule
    \multirow{2}{*}{Pick}
      & OpenVLA-OFT  & 88.0 & 12.0 & 54.0 & 82.0 & 14.0 & 86.0 \\
      & $\pi_{0.5}$  & 84.0 & 10.0 & 62.0 & 86.0 & 22.0 & 80.0 \\
    \midrule
    \multirow{2}{*}{Pick-and-Place}
      & OpenVLA-OFT  & 74.0 &  6.0 & 34.0 & 74.0 & 16.0 & 70.0 \\
      & $\pi_{0.5}$  & 80.0 & 16.0 & 56.0 & 76.0 & 32.0 & 82.0 \\
    \bottomrule
  \end{tabular}
  \caption{Franka Research 3 task success (\%), 50 rollouts per condition. Each model is separately fine-tuned for the robot study; these are within-robot precision comparisons.}
  \label{tab:exp4}

\end{table}

%% file: tex_files/limitation.tex
\section{Limitations}
\label{sec:limitations}
The four evaluated models and their specific checkpoints do not span VLA architectures, and only X-VLA covers benchmarks beyond LIBERO. Some low-precision settings have smaller episode budgets. End-to-end RTN targets eligible linear layers and uses simulated quantization. Real-kernel profiling covers one backbone and GPU. VLA-specific PTQ methods are not compared under matched implementations.

The non-monotone recovery is established for the tested uncalibrated W4A4 RTN scopes. Scope-matched calibration removes the severe joint failures. The findings are not universal properties of four-bit activations. Matched W4/W4A8 controls do not show the severe interaction or numerical recovery. Replay establishes the numerical reversal over 20 full-precision trajectories, but does not isolate its internal cause or reproduce the changed state distribution of a quantized closed-loop policy. The 219 chunks are nested within trajectories, not independent experimental replicates. Small, overlapping diagnostic scope sets support descriptive rank summaries. The separate near-ceiling OpenVLA-OFT trajectory cannot establish a general relationship between action error and task success. Calibration replications cover three two-episode sets from one task and selected scope--strength pairs, rather than a broad calibration search.

The protected OpenVLA-OFT W3 assignment has three evaluation seeds, while the cross-suite protected A8 grid has one per cell. Other quantizers on this failing head remain untested. Fixed tasks and shared initial states limit the binomial approximation. The task-clustered bootstrap uses only ten task clusters and does not preserve episode-level pairing. Seed replication covers selected interventions, and rollout-based timing does not hold observations fixed across methods. The robot study uses separately fine-tuned policies and 50 rollouts per condition, without matched simulation-to-robot controls or physical replication of the nested action-head interventions.

%% file: tex_files/conclusion.tex
\section{Conclusion}
\label{sec:conclusion}
VLAQuantBench shows that quantization behavior depends on the complete precision and calibration recipe. Uncalibrated A4 interventions exhibit numerical and task-level interactions, but two-episode calibration removes the severe failures in the tested $\pi_{0.5}$ action-head subsets. That recipe does not transfer uniformly to $\pi_0$ or OpenVLA-OFT. In OpenVLA-OFT, preserving a single output projection instead recovers near-baseline W3 success on Long and A8 success across four suites. These results support targeted interventions and joint closed-loop evaluation, with the released configurations and records providing a basis for further comparisons.
Evaluating the identified precision assignments (protecting OpenVLA-OFT's output projection; action-head calibration for $\pi_{0.5}$) on physical robots is left to future work and the present robot study reports precision comparisons only.
\label{sec:horizon}

%% file: tex_files/appendix.tex
\section{Quantization Definitions and Implementation}
\label{app:quant-prelim}
\input{figs/overview}
\subsection{Round-to-nearest grids}
\label{appendix:ptq:rtn}
For a tensor group $x$, quantization maps values to an integer grid and immediately dequantizes them:
\begin{equation}
\widehat{x}=s\left[\operatorname{clip}\!\left(\operatorname{round}(x/s)+z,q_{\min},q_{\max}\right)-z\right].
\end{equation}
Here $s$ is the scale and $z$ the zero point. Rounding follows \texttt{torch.round}, including nearest-even tie handling. Grid arithmetic uses float32 and returns to the original tensor dtype. The main accuracy experiments retain floating-point storage and matrix multiplication after this transformation.

For symmetric $b$-bit quantization, $q_{\min}=-2^{b-1}$, $q_{\max}=2^{b-1}-1$, $z=0$, and
\begin{equation}
s=\frac{\max(\max_j |x_j|,10^{-8})}{2^{b-1}-1}.
\end{equation}
For asymmetric quantization, the observed range is extended to include zero. Let $m=\min(\min_j x_j,0)$, $M=\max(\max_j x_j,0)$, and $L=2^b-1$. Then
\begin{equation}
s=\max\!\left(\frac{M-m}{L},10^{-8}\right),\qquad
z=\operatorname{clip}\!\left(\operatorname{round}(-m/s),0,L\right),
\end{equation}
with integer range $[0,L]$. Including zero and bounding the scale handle constant or single-sign groups explicitly.

\subsection{Granularity, scope, and exclusions}
\label{appendix:ptq:setup}
A linear weight matrix has shape $C_{\mathrm{out}}\times C_{\mathrm{in}}$. Per-channel quantization computes one scale per output row. Per-group quantization partitions each row into consecutive groups along the input dimension; a shorter final group gets its own scale without padding. Activations are flattened over leading dimensions and quantized independently along the last dimension, giving one dynamic scale per token.

\begin{table}[htbp]
\centering\small
\begin{tabular}{lcll}
\toprule
Preset & Weight bits & Weight grid & Activation grid \\
\midrule
W3 & 3 & Asymmetric, group 128 & Released precision \\
W4 & 4 & Asymmetric, group 128 & Released precision \\
W8 & 8 & Symmetric, per output channel & Released precision \\
W4A4 & 4 & Asymmetric, group 128 & 4-bit symmetric, per token \\
W4A8 & 4 & Asymmetric, group 128 & 8-bit symmetric, per token \\
W8A8 & 8 & Symmetric, per output channel & 8-bit symmetric, per token \\
\bottomrule
\end{tabular}
\caption{Main RTN presets. W2 uses the W3/W4 weight-grid convention. Additional activation widths and alternative weight grids are restricted to the corresponding ablations.}
\label{tab:quant_presets}
\end{table}

Component scopes select disjoint module roots for VE, MP where present, LLM, and AH. Include/exclude patterns then define layer groups. Only selected linear weights and their inputs are quantized. Biases, embeddings, normalization, attention matrix products, and the language vocabulary head remain at released precision. The X-VLA action decoder implemented through embedding operations is excluded. Thus, ``end-to-end'' denotes all eligible linear scopes, not every operation in the policy. Layer and parameter counts refer to the selected linear weights, not the total model size.

\input{tabs/vla_params}

\subsection{Alternative quantizers}
\label{app:advanced}
AWQ, NF4, LLM.int8(), and SmoothQuant are evaluated on OpenVLA-OFT's LLM scope, while its vision and action modules remain unquantized. These methods differ in their grids, calibration, scaling, and outlier handling; equal nominal bit widths do not make them identical interventions. NF4 evaluates the representation introduced with QLoRA, without QLoRA fine-tuning. The calibration controls on $\pi_{0.5}$ use a separate rollout-based smoothing-and-clipping procedure, detailed in App.~\ref{app:calib}. The real-kernel OpenVLA-OFT configuration uses text calibration.
\FloatBarrier

\section{Benchmark Coverage and Record Selection}
\label{app:benchmarks}
\label{app:protocol_scale}
\paragraph{LIBERO.}
We evaluate the Spatial, Object, Goal, and Long (LIBERO-10) suites, with ten tasks per suite. All four models cover all four suites in the end-to-end grid. X-VLA uses one LIBERO checkpoint across suites. Most configurations use 20 episodes per task. X-VLA normally uses 19 per task under its client, giving 190 episodes; the MLP-fc2 ablation contains 191 distinct episodes and retains that denominator. Policies use official initial states and their own released inference conventions. A cross-model comparison therefore includes checkpoint and client differences, whereas precision comparisons use the corresponding model baseline.

\paragraph{SIMPLER.}
The reported X-VLA visual-matching experiment uses the Google Robot tasks and 864 episodes for each precision setting. To prevent families with more trials from dominating, success is averaged within variants, then across variants within each task family, and finally across families. All seven reported settings have matching variant coverage. The available variant-aggregation runs contain different sets and counts of variants, so their pooled success rates are not used as a controlled precision comparison.

\paragraph{CALVIN.}
The X-VLA experiment uses the $ABC\rightarrow D$ setting. Each evaluation chain contains five sequential subtasks, and the metric is the mean number completed before failure. Six settings use 1,000 chains; W4A4 uses 300. This chain-length metric should not be interpreted as a percentage success rate or assigned a binomial Wilson interval.

\paragraph{VLABench.}
Five evaluation tracks contribute episode progress scores. Baseline, W4, and W8 use 50 episodes per task; W3, W4A8, and W8A8 use 20; W4A4 uses ten. The paper reports the pooled mean progress over the retained episodes, so the weighting follows episode counts. Scenes that fail to compile are omitted consistently across settings. The reduced budgets affect uncertainty and are not hidden by reporting only the mean.

\paragraph{Evaluation seeds.}
The LIBERO runner seeds the simulator, then loads the fixed indexed initial state; changing the seed does not select a different list of initial states. The $\pi$ and X-VLA adapters additionally reset model random-number generators between episodes, affecting stochastic action generation. OpenVLA-OFT uses a deterministic regression head and a fixed model seed (default 7); its evaluation-seed repeats primarily vary simulator random state, not learned weights or sampled action tokens. These are evaluation replications, not independent training runs.

\input{tabs/checkpoints}

\paragraph{Inventory and provenance.}
The analysis manifest lists result-file paths, episode counts, checkpoint identifiers, and SHA-256 hashes. The frozen source list includes the previous 381 runs plus 28 new calibration, scope-split, protected-output, and seed-replication runs; it is not a complete Cartesian product of every model, benchmark, scope, and precision. Separate precision-search experiments and the broader 387-layer exclusion trial are excluded. The 20 reference trajectories collected for $\pi_{0.5}$ replay are listed separately and are not added to the 94,574 closed-loop evaluation episodes. Duplicate task/episode/seed keys are checked within each file. This does not establish independence across files: configurations may share initial states, and layer scopes may overlap. The analyzed inventory comprises 409 evaluation runs and 94,574 simulation episodes. Each run corresponds to one result file; this count includes baseline runs and evaluation-seed repetitions, not 409 distinct precision assignments.

\section{Calibration Details}
\label{app:calib}
The rollout-based control collects statistics at target linear inputs during two unquantized closed-loop episodes. For channel $j$, let $a_j$ be the running maximum of activation magnitude and $w_j=\max_i|W_{ij}|$. After bounding both by $10^{-5}$, the smoothing vector is
\begin{equation}
d_j=\max\!\left(a_j^{\alpha}/w_j^{1-\alpha},10^{-5}\right).
\end{equation}
Weights become $W\operatorname{diag}(d)$ before weight quantization, while the input hook divides activations by $d$. Without clipping or quantization, these inverse transforms preserve the original linear output.

The observer also computes each channel's 99.9th percentile separately for every call and retains the maximum of these per-call quantiles. This is not the percentile of all concatenated calibration tokens. Calls with a single token use its magnitude; exceptionally large inputs are sampled at a fixed stride for quantile computation. The post-smoothing clipping threshold is the retained quantile divided by $d_j$, bounded below by $10^{-5}$. Clipping generally changes the floating-point function and is not part of the exact scaling identity. The dynamic per-token activation quantizer operates after division and clipping.

\paragraph{Calibration episodes and provenance.} The cells in Table~\ref{tab:c3} collected statistics on init states 0--1 of task 0, which are also evaluation states; the replication in Table~\ref{tab:c3_stability} collects on init states 20--25, disjoint from evaluation. The default first calibration state is 20; the statistics cache is keyed on the calibration states, task count, RNG seed, and code revision to distinguish these recorded collection conditions, and every run header records the calibration components, episode range, seed, $\alpha$, clip quantile, cache tag, and code revision.

The real-kernel OpenVLA-OFT SmoothQuant experiment instead uses the method implementation's text calibration with 512 samples and $\alpha=0.85$. AWQ uses 128 calibration samples and group size 128. Their real-kernel settings are separate from the two-episode robot calibration controls and from the RTN accuracy grid.

\paragraph{Model-specific coverage in the new controls.}
New cells collect task-0 states 20--21 at $\alpha=0.5$ unless another strength is stated. Reused $\pi_{0.5}$ LLM/AH/end-to-end reference cells retain their original states 0--1; the separate stability study addresses this overlap. Both $\pi$ models leave MP uncalibrated. OpenVLA-OFT end-to-end calibration includes all five MP layers, all 224 LLM layers, all four AH layers, and 204 of the 209 quantized VE layers observed during calibration. OpenVLA-OFT VE-only coverage is likewise 204/209. Thus, the smoothing/clipping recipe is shared, but observed module coverage and calibration data are explicitly distinguished.
\input{tabs/c3_calibration}
\input{tabs/c3_stability}
\input{tabs/pi0_calibration}

\FloatBarrier
\section{Replication and Uncertainty}
\label{app:replication}
\input{tabs/replication}
For $k$ successful episodes out of $n$, let $\widehat p=k/n$. The reported 95\% Wilson interval uses $z=1.96$:
\begin{equation}
\frac{\widehat p+z^2/(2n)\ \pm\ z\sqrt{\widehat p(1-\widehat p)/n+z^2/(4n^2)}}{1+z^2/n}.
\end{equation}
Intervals describe marginal episode-sampling uncertainty under a binomial approximation. Fixed task allocations, shared initial states, and within-task dependence can limit this approximation. They do not capture training-seed or task-distribution uncertainty. Intervals on individual rates are distinct from intervals on contrasts. Table~\ref{tab:clustered} uses a two-stage percentile bootstrap (5,000 draws, random seed 0): resample ten task identities with replacement, using the same identities for both cells, then independently resample each cell's episodes within the selected task. The task-wise differences are averaged with equal task weight. This implementation preserves task pairing, not paired initial-state outcomes. Only observed tasks enter the resampling distribution.

The table reports selected evaluation-seed repetitions. The attention+adaRMS union and full-head $\pi_{0.5}$ configurations have three seeds, while several other interventions have two. Table~\ref{tab:lattice_object} additionally repeats the full six-setting isolated/joint/recovery comparison on Object at three seeds. Long provides a union-failure observation, but not the complete six-setting comparison. Threshold tables use point estimates within five or ten pp of baseline and are screening summaries, not demonstrated equivalence regions.
\FloatBarrier

\input{tabs/clustered_ci}
\FloatBarrier

\section{Diagnostic Reanalysis}
\label{app:diagnostics}
\input{tabs/proxy_audit}
The saved diagnostics contain per-layer activation kurtosis, relative output error, relative activation error, and activation dispersion. For each matching action-head ablation, the reanalysis applies the recorded include/exclude patterns to diagnostic layer names and checks the selected count against the quantization record. It averages each diagnostic equally across layers in that scope. This weights layers rather than parameters or tokens. Repeated evaluation seeds of the same configuration are excluded from the correlation calculation. Spearman correlations use average ranks for ties.

The diagnostic cohort is frozen to the 35 scopes listed in its source manifest; the new recovery-decomposition scopes are reported separately in Table~\ref{tab:calibrated_subsets}. These 35 scopes are neither disjoint interventions nor independent samples from a population of models. Two models contribute only four scopes each. The original diagnostic files lack complete collection metadata. A separate recollection on initial states 20--21 records checkpoint revision, seed, initial states, MuJoCo, PyTorch, and GPU for each model; code revision is also recorded for the two $\pi$ models. Reapplying the same group-selection and average-rank procedure preserves the four kurtosis--damage rank correlations. This corroborates those rank comparisons under documented collection conditions; it does not supply the missing historical metadata or establish invariance of every diagnostic. The accompanying records distinguish the original analysis from the recollection.

\section{Deployment Measurement Definitions}
\label{app:kernel}
\input{tabs/efficiency_sessions}
The RTX 4090 comparison uses batch size one, the OpenVLA-OFT Spatial checkpoint, and 20 episodes per method. The LLM scope contains 224 selected linear layers. AWQ uses an INT4 GEMM implementation; NF4 and LLM.int8() use their packed method implementations; SmoothQuant uses a \texttt{torch.\_int\_mm} path with online activation quantization. Other policy modules remain at released precision.

The control-call timer encloses the policy adapter call and excludes simulator stepping. An adapter call can either run the network or retrieve an action from a cached chunk. The inference timer measures new network evaluations that generate chunks. The runner discards the first three control calls and first inference in each episode and synchronizes CUDA around timed regions. For both columns in Table~\ref{tab:efficiency}, an episode contributes its recorded mean latency, and the reported value is the median over episodes. These are different estimands from median individual-call latency or total wall-clock time per successful task.

Peak allocated CUDA memory is reported in GiB, with one GiB equal to $2^{30}$ bytes. It excludes nonallocated device memory and is not the checkpoint file size. Actual kernels and packed representations are required for this comparison: simulated RTN keeps floating-point storage. Three additional sessions repeat the rollout-based measurement with 20 episodes per method and show that every quantized path remains slower than BF16; AWQ and SmoothQuant exchange order in control-call latency. Rollout trajectories can differ across methods; these repetitions are not a fixed-observation timing replay. The paired-observation action-fidelity experiment in App.~\ref{app:fidelity} evaluates action errors, not kernel latency. Accordingly, the results describe the measured implementations and workload rather than a universal ordering of quantization algorithms.

\section{Action-Fidelity Measurement}
\label{app:fidelity}
\input{tabs/methods}
\input{tabs/fidelity}
\input{tabs/fidelity_dimensions}
The paired-observation test of Table~\ref{tab:fidelity} rolls out the full-precision policy once on LIBERO-Spatial task 0 from init state 20 (outside the evaluation states 0--19), recording every observation the policy received (all camera images, the proprioceptive state, and the instruction) together with the action it returned. Each quantized policy is then reset for the same task and fed the recorded observations in order, open-loop, so that every policy sees identical inputs; the quantization is applied by the same code path as the closed-loop runs. The reported quantity is the mean over the 80 steps and the seven action dimensions of the absolute difference from the recorded reference actions, in the environment's unnormalized action units. The full-precision replay gives zero measured deviation on this trajectory, checking consistency of this recording and replay. The recorded observations, replayed actions, and per-dimension deviations are retained with the experiment records. Table~\ref{tab:fidelity_dimensions} additionally reports signed mean error and mean absolute error per coordinate over the same 80 control calls. Signed averages can reveal directional bias on this trajectory, but do not establish its accumulation under closed-loop feedback. A single trajectory does not support a population-level estimate of action fidelity or its relationship to task success.

\section{Physical-Robot Setup}
\label{app:physical_robot}
\subsection{Hardware and control}
\label{app:robot_hardware}
\label{app:robot_software}
The physical study uses a Franka Research 3 with a Franka Hand gripper on a tabletop workstation (Figure~\ref{fig:robot_setup}). Two Intel RealSense D415 cameras provide frontal and wrist RGB views; depth is not used. An RTX 4090 workstation runs policy inference, and a separate control computer runs the Polymetis interface to the robot. The machines communicate over Ethernet.
\input{figs/real_robot}
The collection and control software follows DROID~\citep{khazatsky2024droid}. Demonstrations use VR teleoperation with a Meta Quest 3 controller; the controller pose specifies end-effector motion and the trigger controls the gripper. The policy control rate is 15 Hz. Predicted action chunks and the number of actions executed before re-observation are distinct settings and must be recorded with the deployed client.

\subsection{Fine-tuning and evaluation}
OpenVLA-OFT and $\pi_{0.5}$ are each LoRA-fine-tuned using 200 demonstrations. Tables~\ref{tab:pi05_lora_hparams} and~\ref{tab:openvla_oft_lora_hparams} report the study's training configurations. These are robot-adapted policies, not the unchanged simulation checkpoints. The OpenVLA-OFT robot configuration uses two image inputs, proprioception, and a ten-action prediction chunk; its simulation client uses a different chunk convention.
For deployment, OpenVLA-OFT follows its official fine-tuning script, which merges the LoRA adapter into the base weights before saving, so quantization acts on merged weights; $\pi_{0.5}$ uses LeRobot's PEFT loading path, which attaches the adapter without merging.
\input{tabs/app_finetune_params}
Each model is evaluated on Pick and Pick-and-Place at six precision settings, with 50 rollouts per condition. Thus the table contains 1,200 physical rollouts. These comparisons measure precision changes within the physical study. They do not identify the effect of transferring a fixed policy from simulation to reality. The images below illustrate successful and unsuccessful executions; individual examples do not establish mechanisms of quantization failure.
\FloatBarrier

\section{Implementation Checks}
\label{app:canaries}
Reliable interpretation requires recording both the intended intervention and its executed behavior. The checks used in this study concern (i) per-task baseline outcomes under the pinned MuJoCo version; (ii) predicted and executed action-chunk lengths; and (iii) selected linear modules and activation-hook coverage. Layer counts alone cannot detect hooks that were installed but never executed. Known pre-fix results are kept separate from the analyzed corpus.

The evaluation records pair the configuration and checkpoint identifier with simulator versions, seeds, per-episode outcomes, selected module names, and actual activation-hook coverage. Calibration records additionally retain the observation or episode identifiers and cache provenance. The analysis manifest and scripts provide count, aggregation, and source-hash checks. The public repository also supplies environment setup scripts and model-adapter documentation.

\FloatBarrier
\section{Descriptive Precision Screening}
\label{app:screening}
The following summaries apply 10-pp and 5-pp cutoffs to observed point estimates. They describe the tested grid, not confidence-qualified precision budgets. Configurations close to a cutoff require further evaluation before deployment.
\input{tabs/floors}
\input{tabs/budgets}
\FloatBarrier
\section{Complete Action-Head Interventions}
\label{app:interactions}
\input{tabs/lattice}
\input{tabs/lattice_object}
\input{tabs/interaction_controls}
Table~\ref{tab:interaction_controls} collects the additional modulation, activation-width, suite, and flow-step controls referenced in the main text. Its source manifest records the underlying result file and episode denominator for each entry.

\FloatBarrier
\section{Matched-Format Replay and Numerical Diagnostics}
\label{app:pi05_replay}
\paragraph{Records and matched inputs.}
We record a full-precision $\pi_{0.5}$ rollout from initial states 20 and 21 for each of LIBERO-Spatial's ten tasks. These states are disjoint from evaluation states 0--19. All 20 reference rollouts succeed and supply 219 replanning calls, with ten executed actions between replans. Each call predicts a normalized 50-step chunk through ten flow updates. Replay resets the policy per trajectory, restores its recorded random seed, and supplies exactly the recorded images, proprioceptive states, and instruction. Each configuration is loaded separately and uses the same quantization path and checked layer count as closed-loop evaluation. Two full-precision replays reproduce the recorded actions exactly. Quantized policies do not drive the simulator during replay.

\paragraph{Error definitions and aggregation.}
Let $c_{Q,j}$ be predicted normalized chunk $j$ under scope $Q$, and $e_{Q,j}=\operatorname{vec}(c_{Q,j}-c_{\varnothing,j})$. Chunk MAE averages absolute coordinate errors within each full chunk and then over all 219 chunks. Norm ratios, residuals, and cosines are computed per chunk and summarized by medians and interquartile ranges (Table~\ref{tab:replay_composition}). Executed-action MAE and signed bias use the seven environment-action coordinates and pool recorded control calls, including cached actions; coordinate-wise values are supplied separately. Longer trajectories therefore receive more weight in these pooled summaries. Chunks are nested in 20 trajectories and are not treated as 219 independent statistical replicates.

\paragraph{Flow states and input statistics.}
The recorded $x_t$ is the state entering each denoising step, so the tenth-state deviation precedes the final update and differs from final chunk MAE. The state can include padded action coordinates, while returned chunks use the seven action coordinates. Down-projection hooks observe raw inputs before their activation quantizer. Each call records token-wise absmax statistics and the fourth standardized moment of absolute activation values. These summaries are averaged over calls within each trajectory, then over trajectories and the 18 expert down projections. This absolute-value kurtosis differs from a signed-activation statistic. The aggregates can miss particular layers, tokens, or flow steps, and do not establish an outlier-clipping mechanism.

\input{tabs/subsets_formats_pi05}
\input{tabs/replay_composition}
\input{tabs/replay_all_formats}
\input{tabs/replay_bias_diagnostics}

%% file: figs/overview.tex
\begin{figure}[t]
\centering
\includegraphics[width=\linewidth]{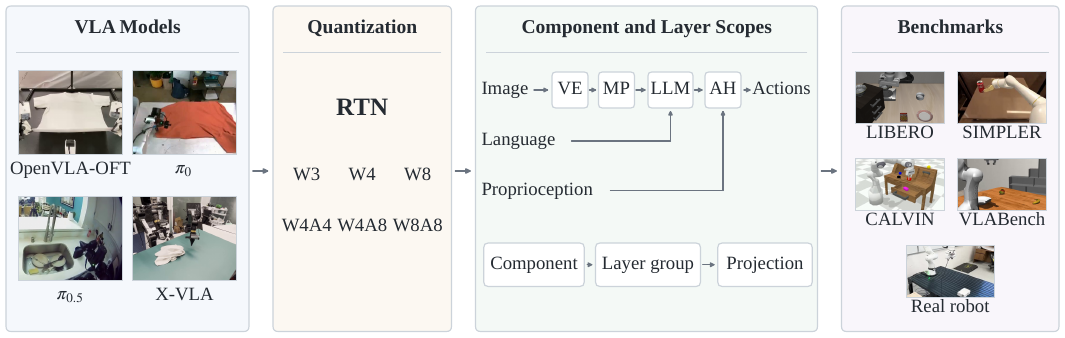}
\caption{Evaluation axes of VLAQuantBench. RTN anchors comparisons across numerical formats and layer scopes. The physical-robot study uses separate fine-tuned policies. Connections summarize component scopes rather than a single architecture shared by all models.}
\label{fig:overview}
\end{figure}

%% file: tabs/vla_params.tex
\begin{table}[htbp]
\centering\small
\begin{tabular}{lrrrr}
\toprule
Model & VE & MP & LLM & AH \\
\midrule
$\pi_{0.5}$ & 162 / 411.1 & 1 / 2.4 & 126 / 1981.8 & 167 / 430.0 \\
$\pi_0$ & 162 / 411.1 & 1 / 2.4 & 126 / 1981.8 & 131 / 314.7 \\
OpenVLA-OFT & 209 / 728.3 & 5 / 88.2 & 224 / 6476.0 & 4 / 151.0 \\
X-VLA & 96 / 335.0 & 0 / 0.0 & 72 / 151.0 & 98 / 304.1 \\
\bottomrule
\end{tabular}
\caption{Selected linear-layer counts / weight parameters in millions for the LIBERO-Spatial W4 end-to-end scopes. Excluded modules are not counted. These are quantized linear weights, not total model parameters or memory measurements. Zero denotes an empty separate scope.}
\label{tab:vla_params}
\end{table}

%% file: tabs/checkpoints.tex
\begin{table}[htbp]
\centering\small
\begin{tabular}{ll}
\toprule
Model / setting & Checkpoint identifier \\
\midrule
OFT / Spatial & \path{moojink/openvla-7b-oft-finetuned-libero-spatial} \\
OFT / Object & \path{moojink/openvla-7b-oft-finetuned-libero-object} \\
OFT / Goal & \path{moojink/openvla-7b-oft-finetuned-libero-goal} \\
OFT / Long & \path{moojink/openvla-7b-oft-finetuned-libero-10} \\
$\pi_0$ / LIBERO & \path{lerobot/pi0_libero_finetuned_v044} \\
$\pi_{0.5}$ / LIBERO & \path{lerobot/pi05_libero_finetuned_v044} \\
X-VLA / LIBERO & \path{2toINF/X-VLA-Libero} \\
X-VLA / SIMPLER-VM & \path{2toINF/X-VLA-Google-Robot} \\
X-VLA / CALVIN & \path{2toINF/X-VLA-Calvin-ABC_D} \\
X-VLA / VLABench & \path{2toINF/X-VLA-VLABench} \\
\midrule
\multicolumn{2}{l}{\textbf{Recollection checkpoints: recorded revision}} \\
OFT / Spatial & \texttt{6d0231af0e48c5985f1ff86908f4674b84bc049b} \\
$\pi_0$ / LIBERO & \texttt{45dcc8fc0e02601c8ccf0554fbd1d26a55070c1f} \\
$\pi_{0.5}$ / LIBERO & \texttt{8e174154ef5f6c60a8da12ae99c303d8963138c1} \\
X-VLA / LIBERO & \texttt{129e71460678b7236cee6fc9707f09d9fa0c3590} \\
\bottomrule
\end{tabular}
\caption{Checkpoint identifiers recorded for the main simulation sweeps. Each $\pi$ checkpoint and the X-VLA LIBERO checkpoint are shared across their four suites. The lower block gives revisions recorded by the four LIBERO diagnostic recollections.}
\label{tab:checkpoints}
\end{table}

%% file: tabs/c3_calibration.tex
\begin{table}[htbp]
  \centering\small
  \begin{tabular}{llc}
    \toprule
    \textbf{Quantized scope} & \textbf{Calibrated} & \textbf{SR \%} \\
    \midrule
    AH only & none & 70.5 \\
     & AH & 98.5 \\
    \midrule
    LLM only & none & 0.0 \\
     & LLM & 70.5 \\
    \midrule
    end-to-end & none & 0.0 \\
     & AH & 0.0 \\
     & LLM & 0.0 \\
     & VE+LLM+AH & 0.0 \\
     & VE+LLM+AH, $\alpha{=}0.75$ & 91.5 \\
    \bottomrule
  \end{tabular}
  \caption{Calibration scope and strength for $\pi_{0.5}$-Spatial at W4A4 (200 evaluation episodes). Statistics use two unquantized-policy episodes, with smoothing strength $\alpha=0.5$ unless noted and 99.9th-percentile clipping. VE+LLM+AH excludes the quantized projector. These are calibration controls, not an optimized SmoothQuant comparison.}
  \label{tab:c3}
\end{table}

%% file: tabs/c3_stability.tex
\begin{table}[htbp]
  \centering\small
  \setlength{\tabcolsep}{5pt}
  \begin{tabular}{lcccc}
    \toprule
    & \multicolumn{4}{c}{\textbf{calibration init states (task 0)}} \\
    \cmidrule(lr){2-5}
    \textbf{Configuration} & \textbf{0--1} & \textbf{20--21} & \textbf{22--23} & \textbf{24--25} \\
    \midrule
    AH only, $\alpha{=}0.5$ & 98.5 & 97.0 & 97.0 & 97.0 \\
    end-to-end VE+LLM+AH, $\alpha{=}0.75$ & 91.5 & 90.5 & 88.0 & 90.5 \\
    \bottomrule
  \end{tabular}
  \caption{Calibration-set stability for $\pi_{0.5}$-Spatial at W4A4 (SR \%, 200 evaluation episodes on init states 0--19 per cell). Each column calibrates on a different two-episode set; the 0--1 column is the original run, whose calibration episodes overlap the evaluation states, and the three others are disjoint from them and from each other. The 95\% Wilson intervals for the disjoint sets are: AH only, [93.6, 98.6] for each set; end-to-end, [85.6, 93.8], [82.8, 91.8], [85.6, 93.8].}
  \label{tab:c3_stability}
\end{table}

%% file: tabs/pi0_calibration.tex
\begin{table}[t]
\centering
\small
\begin{tabular}{lrrrr}
\toprule
Quantized scope & RTN W4A4 & calibrated $\alpha$=0.25 & $\alpha$=0.5 & $\alpha$=0.75 \\
\midrule
VE only & 76.5 & -- & 72.0 & -- \\
LLM only & 72.0 & -- & 11.0 & -- \\
AH only & 58.0 & -- & 39.5 & -- \\
end-to-end & 61.0 & 9.0 & 10.5 & 8.0 \\
\bottomrule
\end{tabular}
\caption{$\pi_0$-Spatial W4A4 calibration controls (SR \%, 200 episodes). Two unquantized episodes at task-0 states 20--21 provide statistics; MP is not calibrated. Dashes denote unmeasured settings. Calibration lowers the measured success point estimates at every scope; the smoothing-strength sweep does not recover the end-to-end baseline.}
\label{tab:pi0_calibration}
\end{table}

%% file: tabs/replication.tex
\begin{table}[htbp]
  \centering\small
  \begin{tabular}{lc}
    \toprule
    \textbf{Cell} & \textbf{SR \% across seeds} \\
    \midrule
    $\pi_{0.5}$ attn$+$adaRMS (W4A4) & 1.5 / 4.0 / 2.5 \\
    $\pi_{0.5}$ all-AH (W4A4) & 70.5 / 70.5 / 68.5 \\
    $\pi_{0.5}$ \texttt{down\_proj} (W4A4) & 83.5 / 86.0 \\
    $\pi_{0.5}$ attention (W4A4) & 98.0 / 98.5 \\
    $\pi_{0.5}$ adaRMS (W4A4) & 93.5 / 95.5 \\
    X-VLA attention (W4A4) & 13.2 / 13.7 \\
    X-VLA \texttt{qkv} (W4A4) & 11.6 / 14.7 \\
    OFT \texttt{fc2} (W4A8) & 18.0 / 17.0 \\
    \bottomrule
  \end{tabular}
  \caption{Selected independent evaluation-seed replications on LIBERO-Spatial. Each run uses 200 episodes, or 190 for X-VLA. These repetitions do not vary the training checkpoint or calibration-data selection.}
  \label{tab:replication}
\end{table}

%% file: tabs/clustered_ci.tex
\begingroup\scriptsize\setlength{\tabcolsep}{3pt}
\begin{longtable}{p{0.47\linewidth}rrrp{0.20\linewidth}}
\caption{Task-clustered uncertainty for selected contrasts. A and B are success rates (\%); $\Delta=A-B$ in pp. Intervals are two-stage bootstrap percentiles (5,000 draws): task identities are paired across cells, but episodes within each selected task are resampled independently. The procedure does not preserve paired initial-state outcomes.}\label{tab:clustered}\\
\toprule Contrast & A & B & $\Delta$ & 95\% interval \\\midrule\endfirsthead
\toprule Contrast (continued) & A & B & $\Delta$ & 95\% interval \\\midrule\endhead
\bottomrule\endfoot
$\pi_{0.5}$ Spatial W4A4: attention vs. baseline & 98.0 & 99.0 & -1.0 & [-4.0, +1.0] \\
$\pi_{0.5}$ Spatial W4A4: adaRMS vs. baseline & 93.5 & 99.0 & -5.5 & [-11.5, -1.0] \\
$\pi_{0.5}$ Spatial W4A4: attention+adaRMS vs. baseline & 1.5 & 99.0 & -97.5 & [-100.0, -93.0] \\
$\pi_{0.5}$ Spatial W4A4: full head (167) vs. 126-layer subset & 70.5 & 7.0 & +63.5 & [+46.0, +80.5] \\
$\pi_{0.5}$ Spatial W4A4: 126 + five conditioning layers (131) vs. 126 & 78.5 & 7.0 & +71.5 & [+51.0, +90.5] \\
$\pi_{0.5}$ Spatial W4A4: 126 + gate/up (162) vs. 126 & 26.5 & 7.0 & +19.5 & [+6.5, +34.5] \\
$\pi_{0.5}$ Spatial W4A4: full head vs. baseline & 70.5 & 99.0 & -28.5 & [-46.5, -11.5] \\
$\pi_{0.5}$ Spatial W4A4, AH-calibrated: attention+adaRMS vs. baseline & 97.5 & 99.0 & -1.5 & [-5.0, +1.5] \\
$\pi_{0.5}$ Spatial W4A4, AH-calibrated: 126-layer subset vs. baseline & 99.0 & 99.0 & +0.0 & [-2.5, +2.5] \\
$\pi_{0.5}$ Spatial W4A4, calibrated end-to-end ($\alpha$=0.75) vs. baseline & 91.5 & 99.0 & -7.5 & [-16.5, +0.0] \\
$\pi_{0.5}$ Object W4A4: full head vs. 126-layer subset & 60.0 & 1.0 & +59.0 & [+37.5, +78.0] \\
X-VLA Spatial W4A4: MLP fc1+fc2 vs. baseline & 24.2 & 96.8 & -72.6 & [-91.6, -49.5] \\
OFT Spatial W4A8: fc2 only vs. residual blocks only & 18.0 & 99.5 & -81.5 & [-92.5, -68.0] \\
OFT Spatial W4A8: end-to-end vs. baseline & 21.5 & 100.0 & -78.5 & [-91.5, -64.5] \\
OFT Spatial W4A8: all but fc2 vs. baseline & 97.5 & 100.0 & -2.5 & [-6.0, +0.0] \\
OFT Object W4A8: all but fc2 vs. baseline & 97.0 & 97.5 & -0.5 & [-5.0, +4.0] \\
OFT Goal W4A8: all but fc2 vs. baseline & 98.0 & 96.0 & +2.0 & [-2.5, +8.0] \\
OFT Long W4A8: all but fc2 vs. baseline & 94.0 & 92.5 & +1.5 & [-6.0, +10.0] \\
OFT Long W3: end-to-end vs. baseline & 8.0 & 92.5 & -84.5 & [-96.5, -66.5] \\
OFT Long W3: fc2 only vs. baseline & 12.5 & 92.5 & -80.0 & [-92.5, -63.5] \\
OFT Long W3: all but fc2 (441 layers) vs. baseline & 93.5 & 92.5 & +1.0 & [-9.5, +11.5] \\
OFT Goal W4: end-to-end vs. baseline & 82.0 & 96.0 & -14.0 & [-27.0, -2.0] \\
$\pi_0$ Spatial W4: end-to-end vs. baseline & 69.0 & 78.0 & -9.0 & [-20.0, +2.5] \\
$\pi_0$ Spatial W4A4: end-to-end vs. baseline & 61.0 & 78.0 & -17.0 & [-31.5, -3.5] \\
$\pi_0$ Spatial W4A4, calibrated end-to-end ($\alpha$=0.5) vs. uncalibrated & 10.5 & 61.0 & -50.5 & [-66.0, -31.5] \\
$\pi_0$ Spatial W4A4, AH-calibrated AH-only vs. uncalibrated AH-only & 39.5 & 58.0 & -18.5 & [-31.5, -5.5] \\
$\pi_0$ Spatial W4A4, LLM-calibrated LLM-only vs. uncalibrated LLM-only & 11.0 & 72.0 & -61.0 & [-75.5, -45.0] \\
$\pi_{0.5}$ Spatial W4A4, 126 + five conditioning layers vs. full head & 78.5 & 70.5 & +8.0 & [-5.0, +23.0] \\
OFT Spatial W4A4, LLM-calibrated LLM-only vs. uncalibrated LLM-only & 33.0 & 0.0 & +33.0 & [+12.0, +55.5] \\
OFT Spatial W4A4, AH-calibrated AH-only vs. uncalibrated AH-only & 38.5 & 0.0 & +38.5 & [+27.5, +49.5] \\
\end{longtable}\endgroup

%% file: tabs/proxy_audit.tex
\begin{table}[htbp]
\centering\small
\begin{tabular}{lrrrrr}
\toprule
Model & Scopes & Kurtosis & Output error & Activation error & Dispersion \\
\midrule
$\pi_{0.5}$ & 19 & +0.35 & +0.31 & +0.46 & -0.04 \\
$\pi_0$ & 4 & -1.00 & -0.40 & -0.40 & -0.80 \\
OpenVLA-OFT & 4 & -0.20 & +0.40 & -0.20 & -0.40 \\
X-VLA & 8 & -0.16 & +0.26 & +0.02 & -0.29 \\
\bottomrule
\end{tabular}
\caption{Exploratory Spearman correlations between saved layer-averaged diagnostics and success loss. Average ranks handle ties; repeated seeds are excluded. Scope sets overlap and per-model samples are small, limiting interpretation; a separately documented recollection reproduces the kurtosis correlations to the reported precision.}
\label{tab:proxy_audit}
\end{table}

%% file: tabs/efficiency_sessions.tex
\begin{table}[htbp]
  \centering\small
  \setlength{\tabcolsep}{3pt}
  \begin{tabular}{lrrrrrrrrrrrr}
    \toprule
    & \multicolumn{4}{c}{Memory (GiB)} & \multicolumn{4}{c}{Control call (ms)} & \multicolumn{4}{c}{Full inference (ms)} \\
    \cmidrule(lr){2-5}\cmidrule(lr){6-9}\cmidrule(lr){10-13}
    Method & ref. & s1 & s2 & s3 & ref. & s1 & s2 & s3 & ref. & s1 & s2 & s3 \\
    \midrule
    BF16 & 14.98 & 14.98 & 14.98 & 14.98 & 18.8 & 18.6 & 18.5 & 18.6 & 139.9 & 138.1 & 137.7 & 137.7 \\
    AWQ & 6.07 & 6.07 & 6.07 & 6.07 & 24.1 & 24.2 & 23.9 & 23.9 & 181.1 & 181.4 & 179.9 & 179.8 \\
    NF4 & 6.31 & 6.31 & 6.31 & 6.31 & 20.2 & 20.3 & 20.4 & 20.7 & 152.6 & 152.2 & 153.0 & 154.4 \\
    LLM.int8() & 8.97 & 8.97 & 8.97 & 8.97 & 48.3 & 47.9 & 47.9 & 47.9 & 379.5 & 376.8 & 375.8 & 376.0 \\
    SmoothQuant & 8.95 & 8.95 & 8.95 & 8.95 & 23.8 & 23.9 & 24.2 & 23.8 & 181.8 & 181.7 & 183.3 & 180.3 \\
    \bottomrule
  \end{tabular}
  \caption{Real-kernel measurements of Table~\ref{tab:efficiency} (ref.) and three further independent profiling sessions (s1--s3), each a fresh process with the five methods interleaved, 20 LIBERO episodes per entry; same GPU and definitions. Across s1--s3, the largest within-method ranges are approximately 0.38\,ms (control call) and 3.00\,ms (full inference); memory variation is below 0.01\,GiB. Ranges are calculated across s1--s3 only.}
  \label{tab:efficiency_sessions}
\end{table}

%% file: tabs/methods.tex
\begin{table}[htbp]
  \centering\small
  \begin{tabular}{llc}
    \toprule
    \textbf{Bits} & \textbf{Method} & \textbf{SR \% [CI]} \\
    \midrule
    \multirow{2}{*}{W3} & RTN & 100.0 [98,100] \\
     & AWQ & 98.5 [96,99] \\
    \midrule
    \multirow{3}{*}{W4} & RTN & 98.0 [95,99] \\
     & AWQ & 99.0 [96,100] \\
     & NF4 & 98.0 [95,99] \\
    \midrule
    \multirow{2}{*}{W8} & RTN & 99.0 [96,100] \\
     & LLM.int8() & 98.5 [96,99] \\
    \midrule
    \multirow{2}{*}{W8A8} & RTN & 100.0 [98,100] \\
     & SmoothQuant & 99.0 [96,100] \\
    \bottomrule
  \end{tabular}
  \caption{LLM-only PTQ on OpenVLA-OFT, LIBERO-Spatial (baseline 100\%; 200 episodes per setting). Brackets give 95\% Wilson intervals. The near-ceiling setting limits resolution of small method differences.}
  \label{tab:methods}
\end{table}

%% file: tabs/fidelity.tex
\begin{table}[htbp]
  \centering\small
  \begin{tabular}{llr}
    \toprule
    \textbf{Bits} & \textbf{Method} & \textbf{mean $|\Delta a|$} \\
    \midrule
    -- & full precision (replay of the reference) & 0.00000 \\
    \midrule
    \multirow{2}{*}{W4} & RTN & 0.00752 \\
     & AWQ & 0.00521 \\
    \midrule
    \multirow{2}{*}{W3} & RTN & 0.01616 \\
     & AWQ & 0.01411 \\
    \bottomrule
  \end{tabular}
  \caption{Action-space fidelity on paired observations (OpenVLA-OFT, LLM-only quantization). One held-out LIBERO-Spatial episode (task 0, init state 20, 80 steps) is rolled out at full precision and its observation sequence is replayed open-loop through each quantized policy; the entry is the mean over steps and the seven action dimensions of the absolute deviation from the reference actions, in the environment's unnormalized action units (App.~\ref{app:fidelity}). The full-precision replay has zero measured deviation on this trajectory.}
  \label{tab:fidelity}
\end{table}

%% file: tabs/fidelity_dimensions.tex
\begin{table}[htbp]
\centering\small\setlength{\tabcolsep}{4pt}
\begin{tabular}{llrrrrrrr}
\toprule
Method & Statistic & $a_1$ & $a_2$ & $a_3$ & $a_4$ & $a_5$ & $a_6$ & $a_7$ \\
\midrule
RTN W4 & Signed & -1.03 & -1.05 & -7.08 & -0.30 & 2.74 & 1.66 & 0.00 \\
RTN W4 & Absolute & 17.90 & 9.57 & 14.96 & 1.71 & 5.45 & 3.06 & 0.00 \\
AWQ W4 & Signed & 1.46 & 2.93 & -1.77 & 0.28 & -0.39 & 0.66 & 0.00 \\
AWQ W4 & Absolute & 9.80 & 10.62 & 10.24 & 1.34 & 2.86 & 1.63 & 0.00 \\
RTN W3 & Signed & -1.14 & -7.49 & -12.37 & -0.15 & 0.75 & 3.90 & 0.00 \\
RTN W3 & Absolute & 48.15 & 18.02 & 27.21 & 4.53 & 9.07 & 6.15 & 0.00 \\
AWQ W3 & Signed & -8.40 & -1.72 & -5.16 & 1.38 & 1.83 & 3.21 & 0.00 \\
AWQ W3 & Absolute & 35.06 & 21.41 & 24.72 & 4.14 & 8.65 & 4.79 & 0.00 \\
\bottomrule
\end{tabular}
\caption{Per-coordinate action deviations on the same held-out OFT trajectory as Table~\ref{tab:fidelity}. Values are multiplied by $10^3$ for readability. Signed error is quantized minus reference action; absolute error is averaged after taking magnitudes. Coordinates follow the client order (three translations, three rotations, gripper); scales are not pooled across coordinates. There are 80 control calls, including actions served from the chunk cache. Full-precision replay is zero in every coordinate.}
\label{tab:fidelity_dimensions}
\end{table}

%% file: figs/real_robot.tex
\begin{figure}[htbp]
    \centering
    \includegraphics[width=0.48\linewidth]{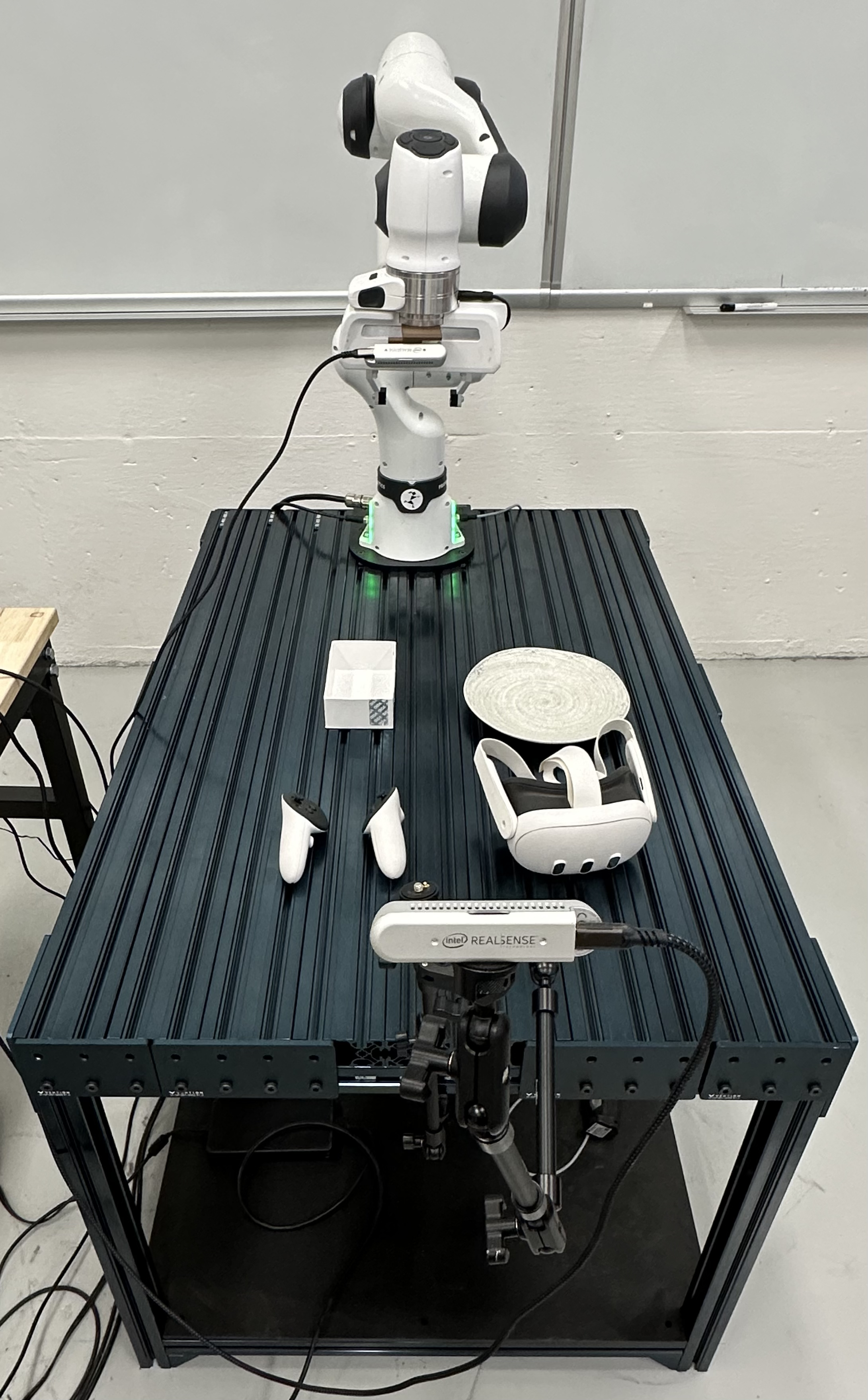}
    \caption{Assembled real-robot platform used in the physical evaluation. 
    The setup consists of the robot arm, end-effector, workspace, 
    and the camera configuration described in this appendix.}
    \label{fig:robot_setup}
\end{figure}

%% file: tabs/app_finetune_params.tex
\begin{table}[htbp]
\centering\small
\caption{Hyperparameters for LoRA fine-tuning of $\pi_{0.5}$.}
\label{tab:pi05_lora_hparams}
\begin{tabular}{ll}
\toprule
Parameter & Value \\
\midrule
\texttt{policy.type} & pi05 \\
\texttt{policy.pretrained\_path} & \texttt{lerobot/pi05\_base} \\
\texttt{policy.freeze\_vision\_encoder} & true \\
\texttt{peft.method\_type} & LORA \\
\texttt{peft.r} & 16 \\
\texttt{policy.optimizer\_lr} & $5\times10^{-5}$ \\
\texttt{batch\_size} & 8 \\
\texttt{steps} & 30{,}000 \\
\texttt{policy.scheduler\_warmup\_steps} & 1{,}000 \\
\texttt{policy.scheduler\_decay\_steps} & 30{,}000 \\
\texttt{policy.scheduler\_decay\_lr} & $5\times10^{-6}$ \\
\texttt{policy.dtype} & bfloat16 \\
\texttt{policy.gradient\_checkpointing} & true \\
\texttt{policy.compile\_model} & false \\
\bottomrule
\end{tabular}
\par\smallskip
\footnotesize Fine-tuned on a single NVIDIA A40. All other hyperparameters follow the LeRobot defaults.
\end{table}

\begin{table}[htbp]
\centering\small
\caption{Hyperparameters for LoRA fine-tuning of OpenVLA-OFT.}
\label{tab:openvla_oft_lora_hparams}
\begin{tabular}{ll}
\toprule
Parameter & Value \\
\midrule
\texttt{vla\_path} & openvla/openvla-7b \\
\texttt{use\_lora} & True \\
\texttt{lora\_rank} & 16 \\
\texttt{use\_l1\_regression} & True \\
\texttt{use\_diffusion} & False \\
\texttt{use\_film} & False \\
\texttt{num\_images\_in\_input} & 2 \\
\texttt{use\_proprio} & True \\
\texttt{learning\_rate} & $5\times10^{-5}$ \\
\texttt{batch\_size} & 8 \\
\texttt{max\_steps} & 100{,}000 \\
\texttt{image\_aug} & True \\
\texttt{NUM\_ACTIONS\_CHUNK}$^\dagger$ & 10 \\
\bottomrule
\end{tabular}
\par\smallskip
\footnotesize Fine-tuned on a single NVIDIA A100-80GB. $^\dagger$Set via the constant in \texttt{prismatic/vla/constants.py}, not a command-line argument. All other hyperparameters follow the official OpenVLA-OFT defaults.
\end{table}

%% file: tabs/floors.tex
\begin{table}[htbp]
  \centering\small
  \setlength{\tabcolsep}{4.5pt}
  \begin{tabular}{lcccccc}
    \toprule
    & \multicolumn{3}{c}{\textbf{Weight-only floor}} & \multicolumn{3}{c}{\textbf{W4A4 (SR \%)}} \\
    \cmidrule(lr){2-4}\cmidrule(lr){5-7}
    \textbf{Model} & VE & LLM & AH & VE & LLM & AH \\
    \midrule
    $\pi_{0.5}$ & 2 & 3 & 3 & \textbf{37.0} & \textbf{0.0} & 70.5 \\
    $\pi_0$ & 2 & 3 & 3 & 76.5 & 72.0 & 58.0 \\
    OpenVLA-OFT & 3 & 3 & 4 & \textbf{0.0} & \textbf{0.0} & \textbf{0.0} \\
    X-VLA & 2 & 2 & 2 & 67.9 & 96.3 & \textbf{0.0} \\
    \bottomrule
  \end{tabular}
  \caption{Component-only quantization on LIBERO-Spatial. The weight-only floor is the lowest tested width among W2/W3/W4 whose success point estimate is within 10 pp of baseline. W4A4 columns quantize only the named component. These thresholds do not establish statistical equivalence.}
  \label{tab:floors}
\end{table}

%% file: tabs/budgets.tex
\begin{table}[htbp]
  \centering\small
  \setlength{\tabcolsep}{4pt}
  \begin{tabular}{llcccccc}
    \toprule
    \textbf{Model} & \textbf{Comp.} & \textbf{A4} & \textbf{A5} & \textbf{A6} & \textbf{A8} & \textbf{none} & \textbf{Budget} \\
    \midrule
    \multirow{3}{*}{$\pi_{0.5}$} & VE & 37.0 & -- & 98.5 & 98.0 & 98.0 & \textbf{A6} \\
     & LLM & 0.0 & 0.0 & 97.5 & 96.0 & 97.5 & \textbf{A6} \\
     & AH & 70.5 & 87.0 & 97.0 & 97.5 & 97.5 & \textbf{A6} \\
    \midrule
    \multirow{3}{*}{$\pi_0$} & VE & 76.5 & -- & 75.5 & 75.0 & 76.5 & \textbf{A4} \\
     & LLM & 72.0 & -- & 72.0 & 76.0 & 76.0 & \textbf{A8} \\
     & AH & 58.0 & -- & 71.5 & 67.0 & 68.5 & \textbf{--} \\
    \midrule
    \multirow{3}{*}{OpenVLA-OFT} & VE & 0.0 & -- & 99.5 & 99.0 & 99.5 & \textbf{A6} \\
     & LLM & 0.0 & -- & 98.0 & 98.5 & 98.0 & \textbf{A6} \\
     & AH & 0.0 & -- & 17.0 & 20.5 & 97.0 & \textbf{none} \\
    \midrule
    \multirow{3}{*}{X-VLA} & VE & 67.9 & -- & 97.9 & 97.4 & 97.4 & \textbf{A6} \\
     & LLM & 96.3 & -- & 97.9 & 96.3 & 95.8 & \textbf{A4} \\
     & AH & 0.0 & 36.8 & 91.1 & 96.8 & 97.4 & \textbf{A8} \\
    \bottomrule
  \end{tabular}
  \caption{Component-only activation precision with W4 weights (LIBERO-Spatial success \%). Budget denotes the lowest tested activation width within 5 pp of the model baseline; none retains released activation precision. Dashes denote unmeasured settings or no qualifying budget. The criterion uses point estimates and does not guarantee joint performance.}
  \label{tab:budgets}
\end{table}

%% file: tabs/lattice.tex
\begin{table}[htbp]
  \centering\small
  \setlength{\tabcolsep}{4.5pt}
  \begin{tabular}{lrrr}
    \toprule
    \textbf{Quantized subset (of the AH)} & \textbf{\#L} & \textbf{SR \% [CI]} & $\Delta$ \\
    \midrule
    attention & 72 & 98.0 [95,99] & -1.0 \\
    adaRMS & 36 & 93.5 [89,96] & -5.5 \\
    attention $+$ adaRMS & 108 & 1.5 [1,4] & -97.5 \\
    \texttt{gate\_proj} & 18 & 99.0 [96,100] & +0.0 \\
    \texttt{up\_proj} & 18 & 98.5 [96,99] & -0.5 \\
    \texttt{down\_proj} & 18 & 83.5 [78,88] & -15.5 \\
    MLP (gate$+$up$+$down) & 54 & 88.0 [83,92] & -11.0 \\
    MLP $+$ adaRMS & 90 & 88.5 [83,92] & -10.5 \\
    \texttt{down}$+$attn & 90 & 80.0 [74,85] & -19.0 \\
    \texttt{down}$+$attn$+$adaRMS & 126 & 7.0 [4,11] & -92.0 \\
    all but adaRMS & 131 & 84.5 [79,89] & -14.5 \\
    \texttt{action\_out\_proj} & 1 & 97.0 [94,99] & -2.0 \\
    all but \texttt{action\_out\_proj} & 166 & 72.0 [65,78] & -27.0 \\
    \textbf{all 167 layers} & 167 & 70.5 [64,76] & -28.5 \\
    \bottomrule
  \end{tabular}
  \caption{Action-head subsets under W4A4 on $\pi_{0.5}$-Spatial (baseline 99.0\%; 200 episodes per setting). Brackets give 95\% Wilson intervals; $\Delta$ is the change from baseline in pp. The 167-layer head contains the 126-layer subset. Selected seed replications appear in App.~\ref{app:replication}.}
  \label{tab:lattice}
\end{table}

%% file: tabs/lattice_object.tex
\begin{table}[htbp]
  \centering\small
  \setlength{\tabcolsep}{5pt}
  \begin{tabular}{lrccc}
    \toprule
    \textbf{Quantized subset (of the AH)} & \textbf{\#L} & \textbf{seed 0} & \textbf{seed 1} & \textbf{seed 2} \\
    \midrule
    full precision & 0 & 99.0 & 98.5 & 99.5 \\
    attention & 72 & 98.5 & 98.5 & 97.5 \\
    adaRMS & 36 & 95.0 & 96.0 & 97.0 \\
    attention $+$ adaRMS & 108 & 2.0 & 2.0 & 2.5 \\
    $+$ \texttt{down\_proj} & 126 & 1.0 & 0.0 & 2.0 \\
    full action head & 167 & 60.0 & 63.0 & 61.0 \\
    \bottomrule
  \end{tabular}
  \caption{The six-setting comparison of Table~\ref{tab:lattice} repeated on LIBERO-Object at three evaluation seeds ($\pi_{0.5}$, W4A4 on subsets of the action head; SR \%, 200 episodes per cell). Non-additivity (the 108-layer union) and recovery (126 to 167 layers) both replicate at every seed.}
  \label{tab:lattice_object}
\end{table}

%% file: tabs/interaction_controls.tex
\begin{table}[htbp]
\centering\small\setlength{\tabcolsep}{4pt}
\begin{tabular}{llllrr}
\toprule
Model & Suite & Format & Intervention & $n$ & SR (\%) \\
\midrule
$\pi_{0.5}$ & Spatial & W4A4 & Attention + pre-attention modulation & 200 & 0.0 \\
$\pi_{0.5}$ & Spatial & W4A4 & Attention + post-attention modulation & 200 & 94.0 \\
$\pi_{0.5}$ & Spatial & W4A4 & q + adaRMS & 200 & 72.0 \\
$\pi_{0.5}$ & Spatial & W4A4 & k + adaRMS & 200 & 88.5 \\
$\pi_{0.5}$ & Spatial & W4A4 & v + adaRMS & 200 & 89.5 \\
$\pi_{0.5}$ & Spatial & W4A4 & o + adaRMS & 200 & 96.5 \\
$\pi_{0.5}$ & Spatial & W4A6 & Attention + adaRMS & 200 & 98.0 \\
$\pi_{0.5}$ & Spatial & W4A8 & Attention + adaRMS & 200 & 98.0 \\
$\pi_{0.5}$ & Long & W4A4 & Attention + adaRMS & 200 & 0.0 \\
X-VLA & Spatial & W4A4 & Both MLP projections & 190 & 24.2 \\
$\pi_{0.5}$ & Spatial & BF16 & Baseline, 1 flow step & 200 & 98.5 \\
$\pi_{0.5}$ & Spatial & W4A4 & Output projection, 1 flow step & 200 & 98.0 \\
$\pi_{0.5}$ & Spatial & BF16 & Baseline, 2 flow steps & 200 & 97.0 \\
$\pi_{0.5}$ & Spatial & W4A4 & Output projection, 2 flow steps & 200 & 99.0 \\
\bottomrule
\end{tabular}
\caption{Supporting scope, format, and flow-step controls. All entries use evaluation seed 0. The source manifest records the exact result files; no new rollouts are introduced by this consolidation.}
\label{tab:interaction_controls}
\end{table}

%% file: tabs/subsets_formats_pi05.tex
\begin{table}[t]
\centering
\small
\begin{tabular}{lrrrr}
\toprule
Quantized subset (of the AH) & \#L & W4 & W4A8 & W4A4 \\
\midrule
full precision & 0 & 99.0 [96,100] & 99.0 [96,100] & 99.0 [96,100] \\
attention & 72 & 100.0 [98,100] & 99.0 [96,100] & 98.0 [95,99] \\
adaRMS & 36 & 99.5 [97,100] & 98.5 [96,99] & 93.5 [89,96] \\
attention + adaRMS & 108 & 99.0 [96,100] & 98.0 [95,99] & 1.5 [1,4] \\
+ down\_proj & 126 & 99.0 [96,100] & 98.5 [96,99] & 7.0 [4,11] \\
full action head & 167 & 97.5 [94,99] & 97.5 [94,99] & 70.5 [64,76] \\
\bottomrule
\end{tabular}
\caption{Matched action-head scopes on $\pi_{0.5}$-Spatial (success \%, 200 episodes per cell; brackets: 95\% Wilson intervals). The six W4/W4A8 settings are within 1.5 pp of the full-precision baseline. W4A4 repeats the reference cells from Table~\ref{tab:lattice}.}
\label{tab:subsets_formats}
\end{table}

%% file: tabs/replay_composition.tex
\begin{table}[htbp]\centering\small
\begin{tabular}{lrrrrr}\toprule
Format & Norm ratio & Residual & Cosine & MAE: 126 / 167 & 167 lower (\%) \\\midrule
W4 & 0.80 & 0.17 & 0.99 & 0.014 / 0.023 & 0.0 \\
W4A8 & 0.80 & 0.20 & 0.98 & 0.015 / 0.024 & 0.0 \\
W4A4 & 2.90 & 0.82 & 0.73 & 0.514 / 0.390 & 99.1 \\
\bottomrule\end{tabular}
\caption{Replay composition statistics over 219 chunks. Norm ratio is $\|e_{A\cup B}\|_2/(\|e_A\|_2+\|e_B\|_2)$; residual is $\|e_{A\cup B}-e_A-e_B\|_2/\|e_{A\cup B}\|_2$; cosine compares $e_{A\cup B}$ with $e_A+e_B$. These columns report medians. MAE is averaged over normalized chunk coordinates and chunks.}
\label{tab:replay_composition}\end{table}

%% file: tabs/replay_all_formats.tex
\begin{table}[htbp]\centering\small\setlength{\tabcolsep}{4pt}
\begin{tabular}{llrrrrr}\toprule
Format & Scope & \#L & SR (\%) & Exec. MAE & Chunk MAE & State MAE \\\midrule
W4A4 & Attention & 72 & 98.0 & 0.0118 & 0.066 & 0.019 \\
 & adaRMS & 36 & 93.5 & 0.0373 & 0.114 & 0.083 \\
 & Union & 108 & 1.5 & 0.1247 & 0.547 & 0.176 \\
 & + down & 126 & 7.0 & 0.1379 & 0.514 & 0.217 \\
 & Full head & 167 & 70.5 & 0.0889 & 0.390 & 0.145 \\
\midrule
W4 & Attention & 72 & 100.0 & 0.0018 & 0.012 & 0.004 \\
 & adaRMS & 36 & 99.5 & 0.0010 & 0.004 & 0.002 \\
 & Union & 108 & 99.0 & 0.0020 & 0.012 & 0.004 \\
 & + down & 126 & 99.0 & 0.0023 & 0.014 & 0.006 \\
 & Full head & 167 & 97.5 & 0.0048 & 0.023 & 0.018 \\
\midrule
W4A8 & Attention & 72 & 99.0 & 0.0018 & 0.012 & 0.004 \\
 & adaRMS & 36 & 98.5 & 0.0011 & 0.004 & 0.002 \\
 & Union & 108 & 98.0 & 0.0020 & 0.012 & 0.004 \\
 & + down & 126 & 98.5 & 0.0027 & 0.015 & 0.006 \\
 & Full head & 167 & 97.5 & 0.0051 & 0.024 & 0.018 \\
\bottomrule\end{tabular}
\caption{Replay across formats on the same 20 held-out trajectories. Executed-action MAE uses environment units; chunk MAE uses the full normalized 50-step output. State MAE is measured before the tenth flow update. SR comes from separate closed-loop evaluations (200 episodes per cell), not the replay trajectories. Definitions and weighting follow App.~\ref{app:pi05_replay}.}
\label{tab:replay_allformats}\end{table}

%% file: tabs/replay_bias_diagnostics.tex
\begin{table}[htbp]\centering\scriptsize\setlength{\tabcolsep}{3pt}
\begin{tabular}{lrrrrrrrrr}\toprule
Scope & $a_1$ & $a_2$ & $a_3$ & $a_4$ & $a_5$ & $a_6$ & $a_7$ & Absmax P99 & $|x|$ kurt. \\\midrule
Attention & -0.005 & -0.002 & 0.004 & 0.000 & 0.000 & -0.000 & 0.005 & 14.99 & 303.0 \\
adaRMS & -0.026 & -0.011 & 0.010 & 0.003 & 0.002 & 0.001 & 0.007 & 15.25 & 213.4 \\
Union & -0.039 & -0.023 & 0.044 & 0.005 & 0.009 & 0.000 & 0.073 & 13.06 & 145.9 \\
126 layers & -0.059 & -0.037 & 0.039 & 0.006 & 0.008 & 0.002 & -0.057 & 14.30 & 157.8 \\
167 layers & -0.032 & -0.017 & 0.008 & 0.002 & 0.005 & 0.001 & -0.004 & 14.18 & 170.9 \\
\bottomrule\end{tabular}
\caption{W4A4 signed action deviations (quantized minus full precision, environment units) and down-projection input summaries. Coordinates follow client order: translation, rotation, gripper. The full-precision input summaries are 15.23 for token-absmax P99 and 315.0 for absolute-value kurtosis. These are averages of per-call statistics, not quantiles of all concatenated tokens.}
\label{tab:replay_bias}\end{table}